\documentclass[10pt,twocolumn,letterpaper]{article}

\usepackage{cvpr}              

\usepackage{graphicx}
\usepackage{verbatim}
\usepackage{amsmath}
\usepackage{amssymb}
\usepackage{booktabs}
\usepackage{pifont}
\newcommand{\cmark}{\ding{51}}%
\newcommand{\xmark}{\ding{55}}%
\usepackage{color}
\usepackage{multirow}
\usepackage{overpic}
\usepackage{arydshln}
\usepackage{makecell}
\usepackage{times}
\usepackage{epsfig}
\usepackage[T1]{fontenc}
\usepackage{hhline}
\usepackage{xcolor}
\usepackage{pifont}
\usepackage{enumitem}
\usepackage{colortbl}
\usepackage{verbatim}
\usepackage{bm}
\usepackage[pagebackref=true,breaklinks=true]{hyperref}

\def\ie{\textit{i.e.}}
\def\eg{\textit{e.g.}}

\definecolor{darkpastelgreen}{rgb}{0.01, 0.75, 0.24}
\definecolor{darkpink}{rgb}{0.91, 0.33, 0.5}
\definecolor{mygray}{gray}{.92}

\definecolor{linkcolor}{RGB}{255,0,0}
\definecolor{urlcolor}{RGB}{255,105,180}
\definecolor{citecolor}{RGB}{0, 80, 200}
\definecolor{citecolor1}{RGB}{0,153,255}
\usepackage{hyperref}
\hypersetup{colorlinks=true,linkcolor=linkcolor,urlcolor=urlcolor,citecolor=citecolor1}

\usepackage[capitalize]{cleveref}
\crefname{section}{Sec.}{Secs.}
\Crefname{section}{Section}{Sections}
\Crefname{table}{Table}{Tables}
\crefname{table}{Tab.}{Tabs.}

\def\cvprPaperID{754} 
\def\confName{CVPR}
\def\confYear{2022}

\title{Learning Affordance Grounding from Exocentric Images}

\begin{document}

\author{Hongchen Luo\textsuperscript{1\rm $\ddagger$}\thanks{This work was done during an internship at JD Explore Academy.} \qquad
Wei Zhai\textsuperscript{1\rm $\ddagger$}\quad Jing Zhang$^3$\thanks{{Corresponding author. $^\ddagger$ Equal contribution.}} \quad Yang Cao$^{1,4}$\footnotemark[2] \quad Dacheng Tao$^{2,3}$
\\
$^{1}$ University of Science and Technology of China \qquad 
\\$^3$ JD Explore Academy \qquad $^2$ The University of Sydney \\
$^4$ Institute of Artificial Intelligence, Hefei Comprehensive National Science Center  \qquad \\
\small \tt \{lhc12,  wzhai056\}@mail.ustc.edu.cn,jing.zhang1@sydney.edu.au, \\
\small \tt forrest@ustc.edu.cn,
dacheng.tao@gmail.com 
}

\maketitle
\begin{abstract}
Affordance grounding, a task to ground (\ie, localize) action possibility region in objects, which faces the challenge of establishing an explicit link with object parts due to the diversity of interactive affordance. Human has the ability that transform the various exocentric interactions to invariant egocentric affordance so as to counter the impact of interactive diversity. To empower an agent with such ability, this paper proposes a task of affordance grounding from exocentric view, \ie, given exocentric human-object interaction and egocentric object images, learning the affordance knowledge of the object and transferring it to the egocentric image using only the affordance label as supervision. To this end, we devise a cross-view knowledge transfer framework that extracts affordance-specific features from exocentric interactions and enhances the perception of affordance regions by preserving affordance correlation. Specifically, an Affordance Invariance Mining module is devised to extract specific clues by minimizing the intra-class differences originated from interaction habits in exocentric images. Besides, an Affordance Co-relation Preserving strategy is presented to perceive and localize affordance by aligning the co-relation matrix of predicted results between the two views. Particularly, an affordance grounding dataset named AGD20K is constructed by collecting and labeling over 20K images from 36 affordance categories. Experimental results demonstrate that our method outperforms the representative models in terms of objective metrics and visual quality. Code:  \href{https://github.com/lhc1224/Cross-view-affordance-grounding}{github.com/lhc1224/Cross-View-AG}.

\end{abstract}

\section{Introduction}
\label{sec:intro}
The goal of affordance grounding is to locate the region of ``action possibilities'' of an object. For an intelligent agent, it is necessary to know not only what the object is but also to understand how it can be used \cite{gibson1977theory}. Perceiving and reasoning about possible interactions in local regions of objects is the key to the shift from passive perception systems to embodied intelligence systems that actively interact with and perceive their environment \cite{bohg2017interactive,nagarajan2019grounded,nagarajan2020learning,ramakrishnan2021exploration}. It has a wide range of applications for robot grasping, scene understanding, action prediction \cite{mandikal2021learning,zhang2020empowering,hassanin2021visual,grabner2011makes,koppula2013learning,luo2021learning,li2021tri,yang2021collaborative}.

\begin{figure}[t]
	\centering
		\begin{overpic}[width=0.92\linewidth]{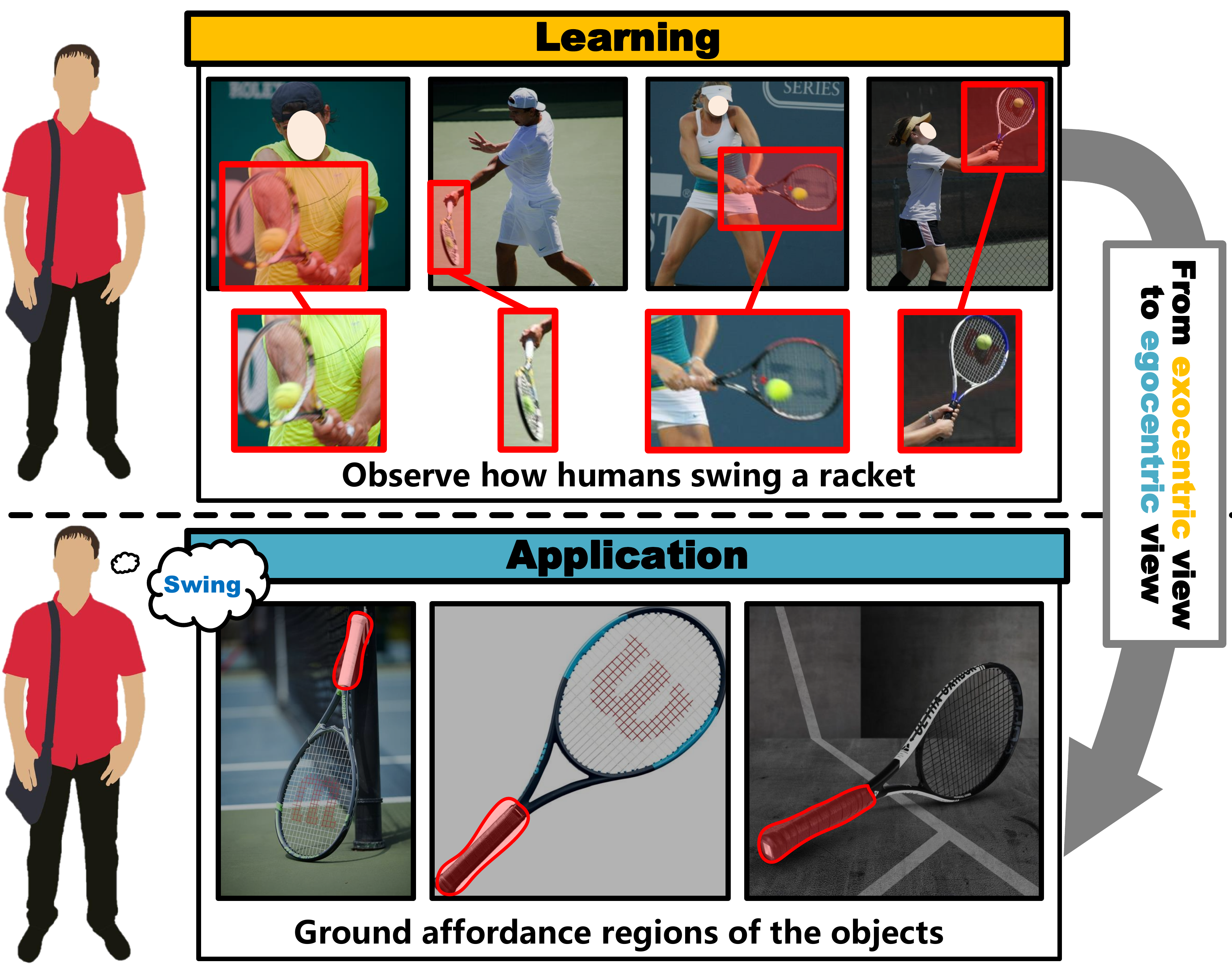}
		    
	\end{overpic}
	\caption{\textbf{Observation.} By observing the exocentric diverse interactions, the human learns affordance knowledge determined by the object's intrinsic properties and transfer it to the egocentric view.}
	\label{figure1}
\end{figure}

\begin{figure*}[t]
	\centering
		\begin{overpic}[width=0.88\linewidth]{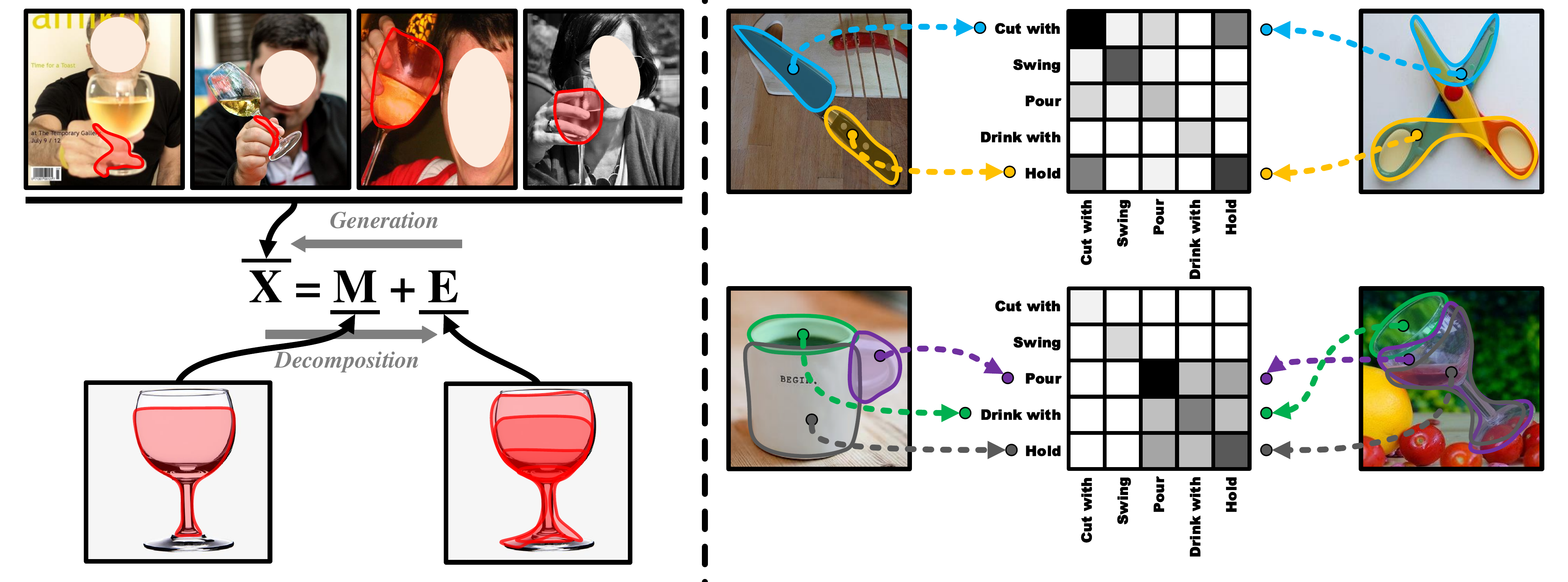}
		\put(21.6,-1){\textbf{(a)}}
		\put(73,-1){\textbf{(b)}}
	\end{overpic}
	\caption{\textbf{Motivation.} 
	 (a) Exocentric interactions can be decomposed into affordance-specific features $M$ and differences in individual habits $E$. (b)  There are co-relations between affordances, \eg ``Cut with'' inevitably accompanies ``Hold'' and is independent of the object category (knife and scissors). Such co-relation is common between objects. In this paper, we mainly consider extracting affordance-specific cues $M$ from diverse interactions while preserving the affordance co-relations to enhance the perceptual capability of the network. 
	}
	\label{motivation}
\end{figure*}

\par As affordance is a dynamic property closely related to the interaction between humans and environment \cite{hassanin2021visual}, it is difficult to understand how to interact with objects and establish an explicit link between the objects' intrinsic properties and affordances \cite{luo2021one}. However, humans can easily perceive the object's affordance region by observing exocentric human-object interactions, and give an egocentric definition. As shown in Fig. \ref{figure1}, although different persons hold the racket in different positions due to their individual habits, the human observer can perceive swingable regions determined by the intrinsic properties (\eg, the long handle structure) of the racket from a group of interacting images, despite the effect of individual differences, and transfer the knowledge to the egocentric view, thereby constructing a bridge between the object part and the affordance category.

\par To empower an agent with this ability to perceive the invariant egocentric affordance from various exocentric interactions, this paper proposes a task of affordance grounding from exocentric view, \ie, given exocentric human-object interactions and egocentric object images, learning affordance knowledge and transferring it to object images by only using affordance labels as supervision. And in the testing stage, the output is the prediction of the affordance region for a specific object with the input of an egocentric object image and a particular affordance label. 
\par To address this problem, we propose a cross-view knowledge transfer framework to extract affordance-specific features from exocentric interactions and transfer them to egocentric view. Specifically, we first devise an Affordance Invariance Mining (AIM) module to decompose the exocentric human-object interactions into the affordance representations determined by objects' intrinsic properties and the differences originated from individual habits (as shown in Fig. \ref{motivation} (a)).  We use low rank matrix decomposition \cite{kolda2009tensor,Lee2000AlgorithmsFN,li2019expectation,geng2021attention} to minimize the intra-class differences caused by diverse interactions to obtain affordance-specific cues. Furthermore, there is a correlation between the object affordances (as shown in Fig. \ref{motivation} (b)), which can be adopted to establish the link between different affordances to reduce the uncertainty caused by multiple affordances regions on the object. Therefore, we present a novel Affordance Co-relation Preserving (ACP) strategy to perceive and localize the affordance region by aligning the co-relation matrix of prediction results from two views. 

Despite the advances in affordance learning, the existing datasets \cite{luo2021one,Sawatzky_2017_CVPR,nguyen2017object,myers2015affordance,fang2018demo2vec} still bear limitations in terms of affordance/object category, image quality, and scene complexity. To carry out a comprehensive study, this paper proposes an affordance grounding dataset named AGD20K, consisting of $20,061$ exocentric images and $3,755$ egocentric images from $36$ affordance categories. The contrastive experiments against several representative methods are performed on the AGD20K dataset. The results demonstrate the superiority of our proposed method in capturing the intrinsic property of objects and suppressing the interactive diversity of affordance.

\noindent\textbf{Contributions:} (1) We present a new affordance grounding from exocentric view task and establish a large-scale AGD20K benchmark to facilitate the research for empowering the agent to capture affordance knowledge from exocentric human-object interactions. (2) We propose a novel cross-view knowledge transfer framework for affordance grounding in which the affordance knowledge is acquired from exocentric human-object interactions and transferred to egocentric views while preserving the correlation between affordances, thereby achieving better perception and localization of interactive affordance. (3) Experiments on the AGD20K dataset demonstrate that our method outperforms state-of-the-art methods and can serve as a strong baseline for future research.

\section{Related Works}
\label{sec:relation work}
\subsection{Visual Affordance Grounding}
The goal of affordance grounding is to locate the region of ``action possibilities'' of an object. Numerous works \cite{nguyen2017object,do2018affordancenet,chuang2018learning,fang2018demo2vec,zhao2020object,koppula2014physically,zhai2021one,lu2022phrase} mainly build upon supervised approaches to establish mapping relations between local regions of objects and affordance. Sawatzky et al. \cite{Sawatzky_2017_CVPR,sawatzky2017adaptive} adopt an Expectation-Maximization algorithm \cite{dempster1977maximum} to achieve weakly supervised affordance detection using only a few key points. Nagarajan et al. \cite{nagarajan2019grounded} exploit only affordance labels to ground the interactions from the videos. In contrast to \cite{nagarajan2019grounded}, our goal is to empower the agent to learn affordance knowledge from exocentric human-object interactions. To this end, we propose an explicit cross-view knowledge transfer framework that extracts affordance knowledge determined by the intrinsic properties of objects from multiple exocentric interactions and transfers it into egocentric images.

\subsection{Visual Affordance Dataset}
The emergence of the relevant datasets drives the development of affordance grounding, as shown in Table \ref{Table:relevant datasets}. For example, Sawatzky et al. \cite{Sawatzky_2017_CVPR} select video frames from CAD120 \cite{koppula2013learning} to construct a weakly supervised affordance detection dataset, using only cropped out object regions but in inferior image quality. Other affordance-related datasets \cite{myers2015affordance,nguyen2017object,chuang2018learning,fang2018demo2vec,roy2016multi} face the problems of small scale and low affordance/object category diversity and do not consider human actions to reason about the affordance regions. PAD dataset \cite{luo2021one} considers the inference of human purpose from support images of human-object interactions and transfers to a group of query images but does not provide part-level affordance labels. In contrast to the above works, we explicitly consider exocentric-to-egocentric viewpoint transformations and collect a much larger scale of images, with richer affordance/object categories and part-level annotations, which are more useful and applicable to real-world application domains.

\subsection{Learning View Transformations}
The existing learning-view transformation works start from the theory of mirror neurons \cite{rizzolatti2004mirror}, which adopts embedding learning to generate perspective invariant representations from paired data, and leverage it for tasks such as action recognition and video summarization under egocentric view \cite{sigurdsson2018actor,soran2014action,ho2018summarizing,regmi2019bridging}. For example, Li et al. \cite{li2021ego} extract key egocentric signals from the exocentric view dataset during pre-training and distill them to the backbone to guide feature learning in the egocentric video task. In contrast to the above works, we aim to extract affordance knowledge from the diverse exocentric human-object interactions and transfer it to the egocentric view, which is challenging due to the uncertainty caused by various interactions and the multiple affordance regions that objects contain.

\section{Method}
\label{sec:method}
Our goal is to ground the object affordance regions in egocentric images. During training, given a group of exocentric images $\mathcal{I}_{exo}=\{I_1,...,I_N\}$ ($N$ is the number of exocentric images) and an egocentric object image $I_{ego}$, the network uses only affordance labels as supervision, so as to learn affordance knowledge from exocentric images and transfer it to egocentric images. During testing, only given an egocentric image $I_{ego}$ and the affordance label $C_a$, the network outputs the affordance region on the object.

\begin{table}[!t]
    \centering
  \scriptsize
  \renewcommand{\arraystretch}{1.}
  \renewcommand{\tabcolsep}{7.2pt}
   \caption{\textbf{Statistics of related datasets and the proposed AGD20K dataset.} Part: part-level annotation. HQ: high-quality annotation. $\sharp$Obj: number of object classes. $\sharp$Aff: number of affordance classes. $\sharp$Img: number of images.}
\label{Table:relevant datasets}
  \begin{tabular}{c|c|cc|cccc}
\hline
\Xhline{2.\arrayrulewidth}
\textbf{Dataset}      & \textbf{Year} & \textbf{HQ}  & \textbf{Part} &  \bm{$\sharp$}\textbf{Obj.} & \bm{$\sharp$}\textbf{Aff.} & \bm{$\sharp$}\textbf{Img.}   \\
\hline
\Xhline{2.\arrayrulewidth}
 UMD \cite{myers2015affordance}        & 2015  & \xmark  & \cmark   & 17  & 7  & 30,000 \\
  \cite{Sawatzky_2017_CVPR} &  2017   & \xmark & \cmark  & 17 & 7 & 3,090  \\
 IIT-AFF \cite{nguyen2017object}     & 2017 & \xmark &  \cmark        & 10  & 9   & 8,835  \\
 ADE-Aff \cite{chuang2018learning}     &  2018   & \cmark   & \cmark    & 150 & 7   & 10,000 \\
 PAD \cite{luo2021one}     & 2021  & \cmark   & \xmark     & 72  & 31  & 4,002  \\
\hline
\rowcolor{mygray}
AGD20k (Ours) &  2021  & \cmark  & \cmark & 50 & 36 & 23,816   \\
\hline
\Xhline{2.\arrayrulewidth}
    \end{tabular}
\end{table}

\newcommand{\tabincell}[2]{\begin{tabular}{@{}#1@{}}#2\end{tabular}}
\begin{figure*}[t]
	\centering
		\begin{overpic}[width=0.94\linewidth]{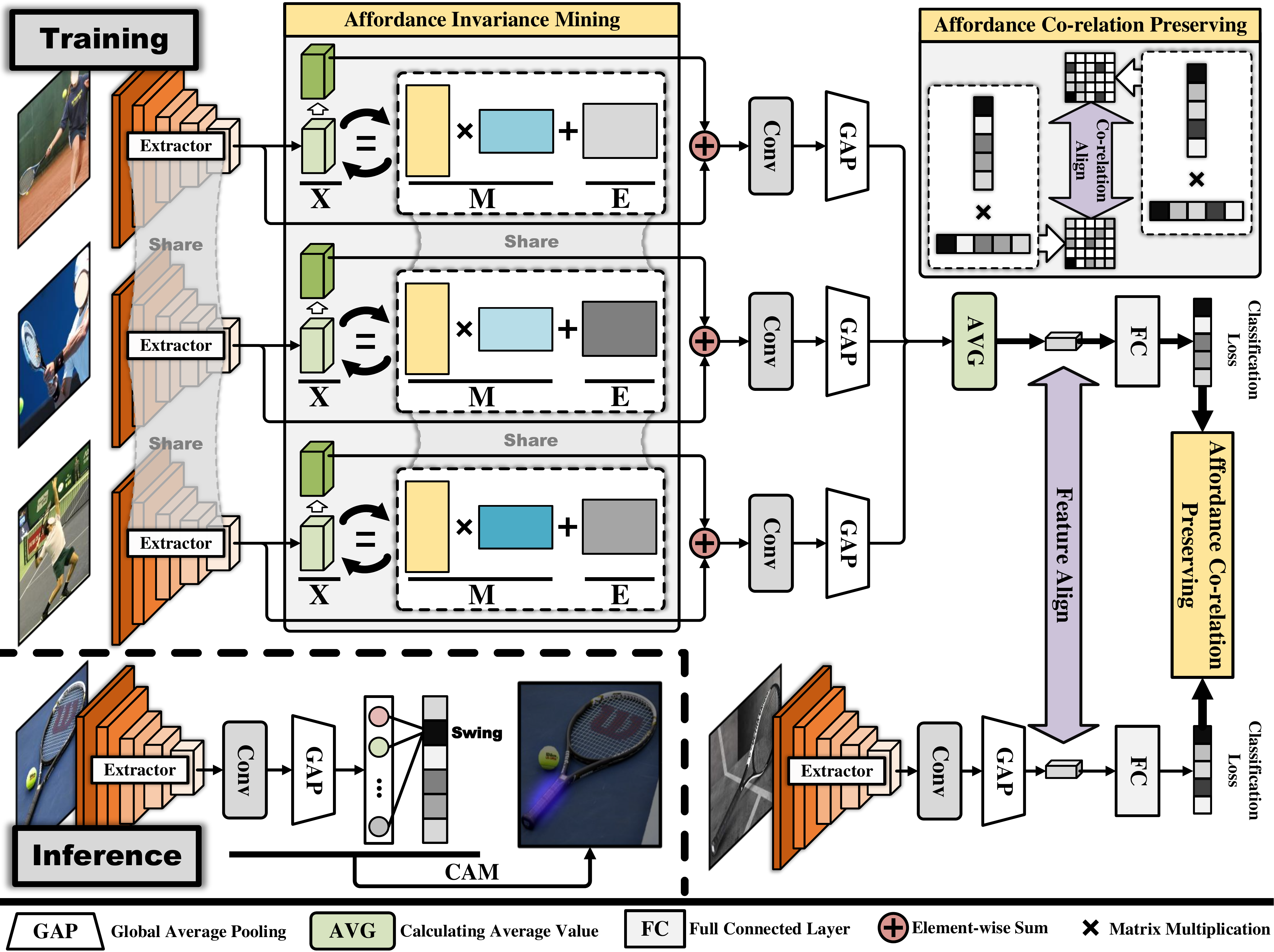}
		\put(26,58.1){\scriptsize{$\bm{1}$}}
		\put(25.9,42.6){\scriptsize{$\bm{2}$}}
		\put(25.9,26.7){\scriptsize{$\bm{N}$}}
		\put(39,27){\scriptsize{$\bm{N}$}}
		\put(49.2,27){\scriptsize{$\bm{N}$}}
		
		\put(39,42.7){\scriptsize{$\bm{2}$}}
		\put(49.3,42.7){\scriptsize{$\bm{2}$}}
		
		\put(39,58.1){\scriptsize{$\bm{1}$}}
		\put(49.3,58.1){\scriptsize{$\bm{1}$}}
		
		\put(32.45,63.7){\small{$\bm{W}$}}
		\put(32.43,48.1){\small{$\bm{W}$}}
		\put(32.4,32.8){\small{$\bm{W}$}}
		
		\put(38.8,32.7){\small{$\bm{H_N}$}}
		\put(39,48.3){\small{$\bm{H_2}$}}
		\put(39.2,63.8){\small{$\bm{H_1}$}}
		
		\put(26.6,68.1){\small{$\bm{M_1}$}}
   		\put(26.6,52.6){\small{$\bm{M_2}$}}
		\put(26.6,37.15){\small{$\bm{M_N}$}}
		
		\put(30,70.4){\small{$\bm{Conv}$}}
		\put(30,54.9){\small{$\bm{Conv}$}}
		\put(30,39.3){\small{$\bm{Conv}$}}
		
		\put(39,5.8){\small{\cite{zhou2016learning}}}
		
		\put(82,11.5){\small{\bm{$f_{ego}$}}}
		\put(82,49.5){\small{\bm{$f_{exo}$}}}
		\put(19.25,64){\small{\bm{$Z_1$}}}
		\put(19.25,48.2){\small{\bm{$Z_2$}}}
		\put(19,33){\small{\bm{$Z_N$}}}
		\put(81.7,68){\scriptsize{\bm{$Q$}}}
		\put(87.7,54.9){\scriptsize{\bm{$P$}}}
		\put(74.3,63.2){\small{\bm{$p$}}}
		\put(95.5,65.8){\small{\bm{$q$}}}
		\put(82.,64.5){\rotatebox{270}{\textbf{\scriptsize{(Eq. \ref{LACP})}}}}
		
		\put(79,41.8){\rotatebox{270}{\small{\bm{$L_{KT}=||f_{exo}-f_{ego}||$}}}}
		\put(89.5,38){\rotatebox{270}{\textbf{\small{(Eq. \ref{acp1}) $ \sim $ (Eq. \ref{LACP})}}}}
	\end{overpic}
	\caption{\textbf{Overview of the proposed cross-view knowledge transfer affordance grounding framework.} It mainly consists of an Affordance Invariance Mining (AIM) module and an Affordance Co-relation Preservation (ACP) strategy. The AIM module (see in Sec. \ref{AIMmodule}) aims to obtain invariant affordance representations from diverse exocentric interactions. The ACP strategy (see in Sec. \ref{ACP}) enhances the network's affordance perception by aligning the co-relation of the outputs of the two views.}
	\label{pipeline}
\end{figure*}
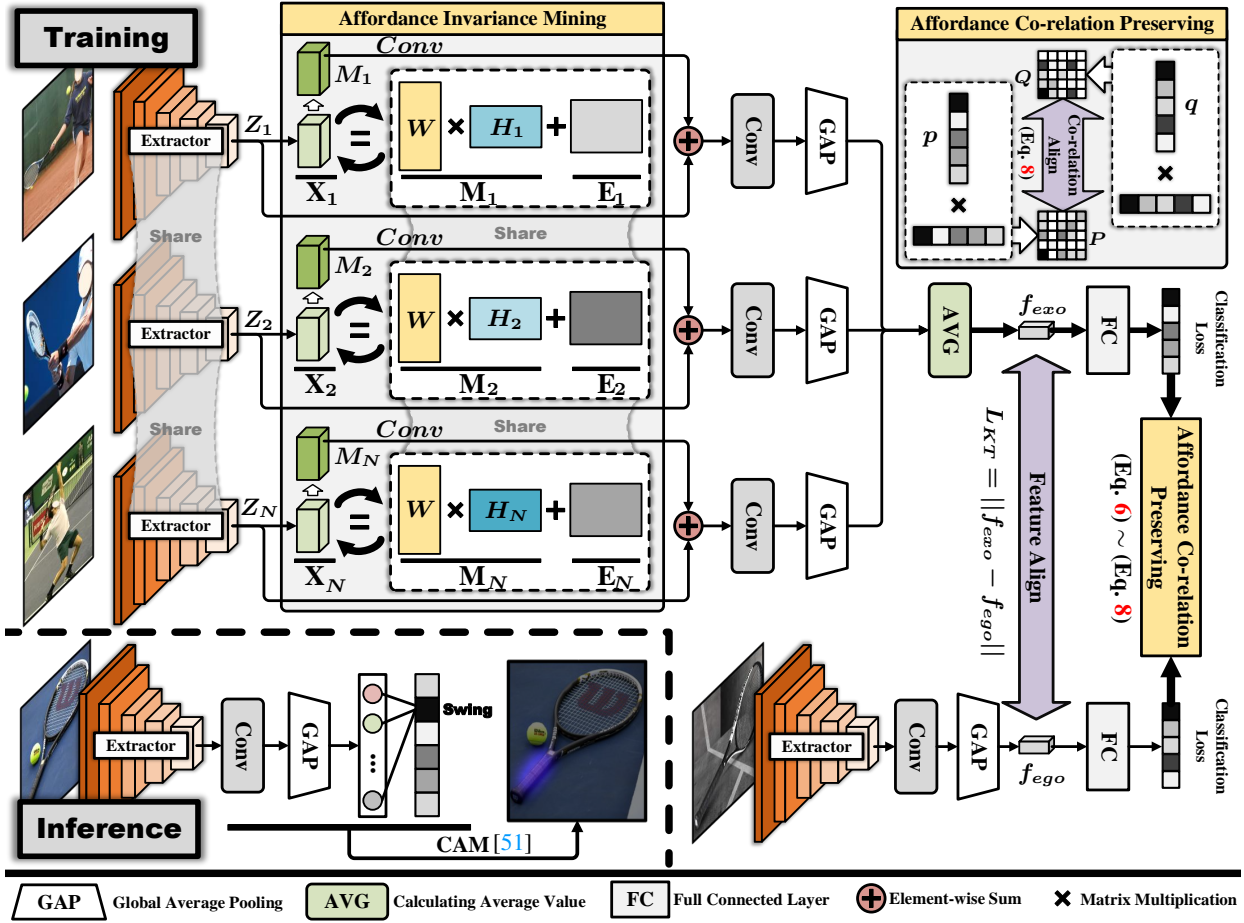

\par Our proposed cross-view knowledge transfer framework for affordance grounding is shown in Fig. \ref{pipeline}. During training, we first use Resnet50 \cite{he2016deep} to extract the features of exocentric and egocentric images to obtain $\mathcal{Z}_{exo}=\{Z_1,...,Z_N\}$ and $Z_{ego}$, respectively. We then present the Affordance Invariance Mining (AIM) module (see in Sec. \ref{AIMmodule}) to extract affordance-specific clues ($\mathcal{F}_{exo}$) from the exocentric features. Meanwhile, we use two convolutional layers to map the egocentric feature to the embedding space consistent with the exocentric view: $F_{ego}=Conv(Z_{ego})$. Subsequently, the features of the two branches ($\mathcal{F}_{exo}$ and $F_{ego}$) are fed into the same convolution layer to obtain features $\mathcal{D}_{exo}$ and $D_{ego}$ respectively. To ensure the affordance knowledge can be transferred to the egocentric view, we average the $\mathcal{D}_{exo}$ through the global average pooling (GAP) layer to obtain the $f_{exo}$ and pass the $D_{ego}$ through the GAP layer to get the $f_{ego}$, and align $f_{exo}$ and $f_{ego}$ using L2 loss $L_{KT}$. Then, $f_{exo}$ and $f_{ego}$ are fed into the same fully connected layer to obtain the affordance prediction. Finally, we propose an Affordance Co-relation Preserving (ACP) strategy (see in Sec. \ref{ACP}) to enhance the network's perception of affordance by aligning the co-relation matrix of the outputs of the two views.  During testing, we feed the egocentric object images into the network only through the egocentric branch, and then use the CAM \cite{zhou2016learning} technique to obtain the affordance regions of the object (see in Sec. \ref{TP}). 

\begin{figure*}[t]
	\centering
		\begin{overpic}[width=0.95\linewidth]{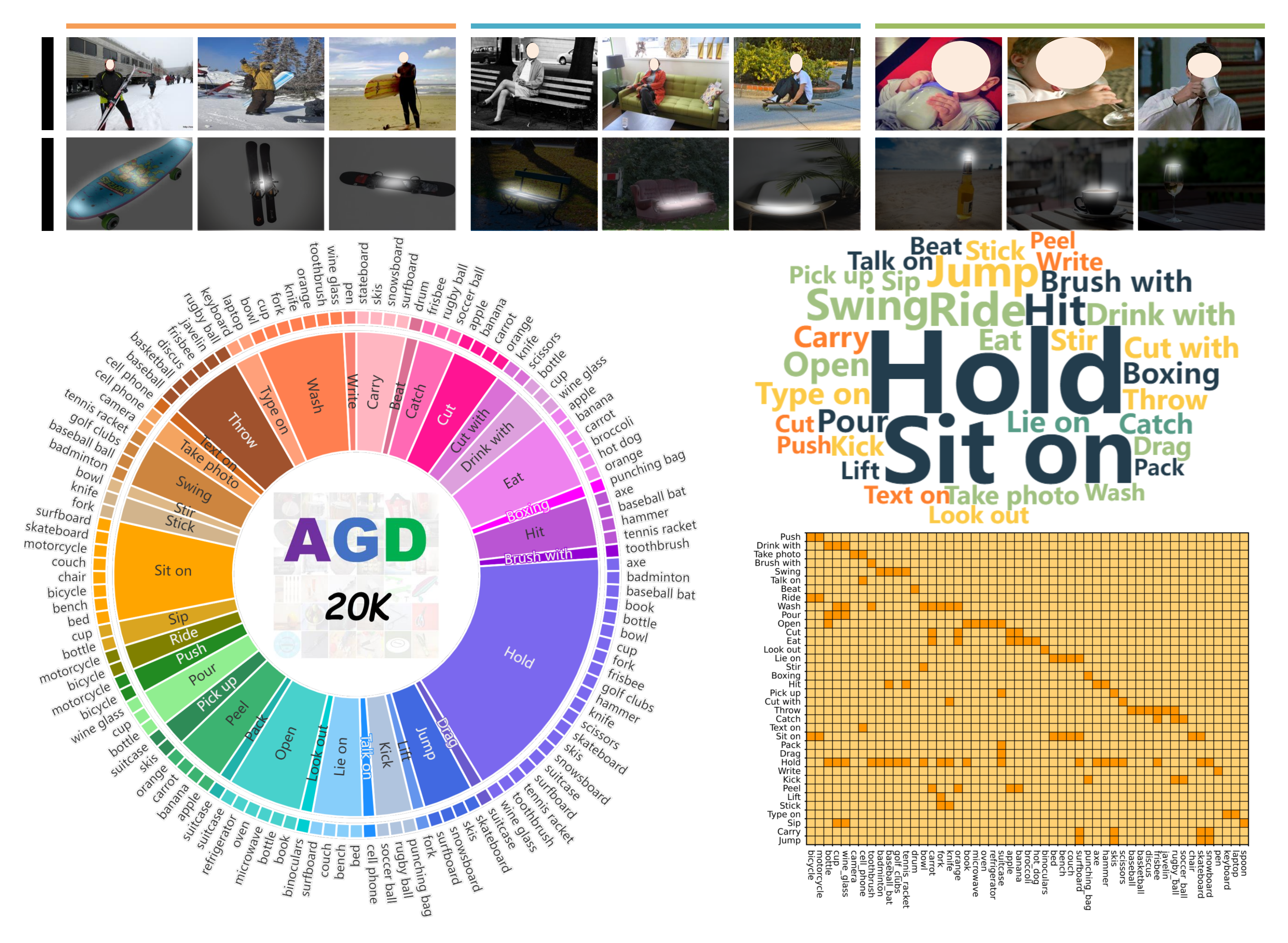}
    		\put(0,72){\colorbox{black}{{\color{white} \textbf{(a)}}}}
            \put(0,4){\colorbox{black}{{\color{white} \textbf{(b)}}}}
            \put(57,52.5){\colorbox{black}{{\color{white} \textbf{(c)}}}}
            \put(57,4){\colorbox{black}{{\color{white} \textbf{(d)}}}}
            
            \put(0,64.5){\colorbox{white}{{\color{black} \rotatebox{90}{\small\textbf{Train}}}}}
            \put(0,56.5){\colorbox{white}{{\color{black} \rotatebox{90}{\small\textbf{Test}}}}}
            
            \put(16,74.5){{\color{black} \small\textbf{Carry}}}
            \put(49,74.5){{\color{black} \small\textbf{Sit on}}}
            \put(79.5,74.5){{\color{black} \small\textbf{Drink with}}}
            
            \put(7,72){{\color{black} \footnotesize\textbf{Skis}}}
            \put(15,72){{\color{black} \footnotesize\textbf{Snowboard}}}
            \put(26,72){{\color{black} \footnotesize\textbf{Surfboard}}}
            \put(39,72){{\color{black} \footnotesize\textbf{Bench}}}
            \put(49,72){{\color{black} \footnotesize\textbf{Couch}}}
            \put(58,72){{\color{black} \footnotesize\textbf{Skateboard}}}
            \put(72,72){{\color{black} \footnotesize\textbf{Bottle}}}
            \put(80.5,72){{\color{black} \footnotesize\textbf{Wine glass}}}
            \put(93,72){{\color{black} \footnotesize\textbf{Cup}}}
            
	\end{overpic}
	\caption{\textbf{The properties of the AGD20K dataset.} (a) Some examples from the dataset. (b) The distribution of categories in AGD20K. (c) The word cloud distribution of affordances in AGD20K. (d) Confusion matrix between the affordance category and the object category in AGD20K, where the horizontal axis denotes the object category and the vertical axis denotes the affordance category. }
	\label{dataset}
\end{figure*}

\subsection{Affordance Invariance Mining Module}
\label{AIMmodule}
As shown in Fig. \ref{pipeline}, we decompose the interactions in exocentric images into affordance-specific features $M$ and individual differences $E$. Inspired by low-rank matrix decomposition \cite{kolda2009tensor,Lee2000AlgorithmsFN,geng2021attention}, we represent the $M$ as the multiplication of a dictionary matrix $W$ and a corresponding matrix $H$, where the dictionary bases represent the sub-features of human-object interaction, and minimize $E$ by iterative optimization to obtain a reconstructed affordance representation $M$.  Specifically, for the input $Z_i$, we first reduce its dimensionality with a convolution layer and a ReLU layer to ensure the non-negativity of the input, and then reshape them into $X_i \in R^{c \times hw}$ ($c$, $h$ and $w$ are the channels, length, and width of the feature maps respectively). We use non-negative matrix factorization (NMF) \cite{Lee2000AlgorithmsFN} to update the dictionary and the coefficient matrices. Consequently, $X_i$ is decomposed into two non-negative matrices $W$ and $H_i$. Here $W \in R^{c \times r}$ is the dictionary matrix shared by all exocentric features, while $H_i \in R^{r \times hw}$ is the coefficient matrix of each exocentric feature, and $r$ is the rank of the low-rank matrix $W$. To update $H_i$ and $W$ in parallel, we concatenate $\mathcal{X}_{exo}=\{X_1,...,X_N\}$ and $\mathcal{H}=\{H_1,...,H_N\}$ to obtain $X \in R^{c \times Nhw}$ and $H \in R^{r \times Nhw}$. Mathematically, the optimization process can be formulated as follows: 
\begin{equation}
\small
    \mathop{\min}\limits_{W,H} ||X-WH||, \quad s.t. \ W_{ab}\ge 0,H_{bk} \ge 0. \label{eq1}
\end{equation}
$W$ and $H$ are updated according to the following rules:
\begin{equation}
\small
    H_{ab} \leftarrow H_{ab}\frac{(W^TX)_{ab}}{(W^TWH)_{ab}}, W_{ab} \leftarrow W_{ab}\frac{(XH^T)_{ab}}{(WHH^T)_{ab}}. \label{WH}
\end{equation}

After several iterations, we get the output $M=WH$, and reshpae it to $\mathcal{M}_{exo}=\{M_1,..., M_N\}, M_i \in R^{c \times h \times w}$. Finally, we use a convolution layer to map it to the residual space and sum it with the $\mathcal{Z}$ to get the final output $\mathcal{F}_{exo}$:
\begin{equation}
\small
      F_i=Z_i+Conv(M_i), \quad i\in [1,N]. \label{eq4}
\end{equation}
In each batch of training, we update the initial dictionary matrix $W^{(0)}$ such that it can contain the statistical prior of the common subfeature of human-object interaction, \ie,
\begin{equation}
\small
    W^{(0)} \leftarrow \alpha W^{(0)}+(1-\alpha)\bar{W}, \label{eq5}
\end{equation}
where $\bar{W}$ is the average over each mini-batch.

\subsection{Affordance Co-relation Preserving Strategy}
\label{ACP}
As shown in Fig. \ref{pipeline}, we feed the feature representations of the two branches ($f_{exo}$ and $f_{ego}$) into the same fully connected layer respectively to obtain the affordance category prediction scores $s$ and $g$:
\begin{equation}
\small
    s=FC(f_{exo}), \quad g=FC(f_{ego}).
\end{equation}
Then, we align the affordance co-relation between the exocentric and egocentric views by calculating the cross-entropy loss \cite{hinton2015distilling}  $L_{ACP}$ of the co-relation matrix of the prediction scores of the two branches:
\begin{equation}
\small
    p_j=\frac{exp(s_j/\boldsymbol{T})}{\sum_k^{N_c}exp(s_k/\boldsymbol{T})},\quad q_j=\frac{exp(g_j/\boldsymbol{T})}{\sum_k^{N_c}exp(g_k/\boldsymbol{T})}, \label{acp1}
\end{equation}
\begin{equation}
\small
    P=pp^T, Q=qq^T, \label{acp2}
\end{equation}
\begin{equation}
\small
    L_{ACP}=-\sum_j^{N_c}\sum_k^{N_c} P_{jk} log(Q_{jk}), \label{LACP} 
\end{equation}
where $\boldsymbol{T}$ is used to control the degree of attention paid to the correlations between negative labels. $P_{jk}$ and $Q_{jk}$ denote the correlation between classes $j$ and $k$ in the prediction results. Finally, the total loss can be calculated as:
\begin{equation}
\small
    L=\lambda_1 L_{cls}+\lambda_2 L_{ACP} + \lambda_3 L_{KT}, \label{totalL}
\end{equation}
where $\lambda_1$, $\lambda_2$ and $\lambda_3$ are hyper-parameters to balance the classification loss, ACP loss and $L_{KT}$ loss. $L_{cls}$ is the sum of the cross-entropy losses of the classification results of the two branches, and $L_{KT}$ is loss of cross-view affordance knowledge transfer: $L_{KT}=||f_{exo}-f_{ego}||$.

\begin{table*}[!t]
    \centering
  \footnotesize
  \renewcommand{\arraystretch}{1.}
  \renewcommand{\tabcolsep}{11.6pt}
   \caption{\textbf{The results of different methods on AGD20k.} The best results are in \textbf{bold}. ``Seen'' means that the training set and the test set contain the same object categories, while ``Unseen'' means that the object categories in the training set and the test set do not overlap. The $\textcolor{darkpink}{\diamond}$ defines the relative improvement of our method over other methods. ``\textcolor[rgb]{0.6,0.1,0.1}{Dark red}'', ``\textcolor[rgb]{0.99,0.5,0.0}{Orange}'' and ``\textcolor[rgb]{0.4,0.0,0.99}{Purple}'' represent saliency detection, weakly supervised object localization and affordance grounding models, respectively.}
   \label{Table:1}
  \begin{tabular}{r||ccc|ccc}
    \hline
    \Xhline{2.\arrayrulewidth}
   & \multicolumn{3}{c|}{\textbf{Seen}} & \multicolumn{3}{c}{\textbf{Unseen}}   \\
   \hline
   \textbf{Methods} & $\text{KLD} \downarrow$ & $\text{SIM} \uparrow$ & $\text{NSS} \uparrow$  & $\text{KLD} \downarrow$ & $\text{SIM} \uparrow$ & $\text{NSS} \uparrow$
\\   \hline
\Xhline{2.\arrayrulewidth}
  \textcolor[rgb]{0.6,0.1,0.1}{Mlnet}  \cite{cornia2016deep} & $5.197\textcolor{darkpink}{\scriptstyle~\diamond70.4\%}$ & $0.280\textcolor{darkpink}{\scriptstyle~\diamond19.3\%}$ & $0.596\textcolor{darkpink}{\scriptstyle~\diamond55.5\%}$ & $5.012\textcolor{darkpink}{\scriptstyle~\diamond64.3\%}$ & $0.263\textcolor{darkpink}{\scriptstyle~\diamond8.4\%}$ & $0.595\textcolor{darkpink}{\scriptstyle~\diamond39.3\%}$\\
  \textcolor[rgb]{0.6,0.1,0.1}{DeepGazeII} \cite{kummerer2016deepgaze} & $1.858\textcolor{darkpink}{\scriptstyle~\diamond17.2\%}$ & $0.280\textcolor{darkpink}{\scriptstyle~\diamond19.3\%}$ & $0.623\textcolor{darkpink}{\scriptstyle~\diamond48.8\%}$ & $1.990\textcolor{darkpink}{\scriptstyle~\diamond10.2\%}$ & $0.256\textcolor{darkpink}{\scriptstyle~\diamond11.3\%}$ & $0.597\textcolor{darkpink}{\scriptstyle~\diamond38.9\%}$ \\
  \textcolor[rgb]{0.6,0.1,0.1}{EgoGaze} \cite{huang2018predicting} & $4.185\textcolor{darkpink}{\scriptstyle~\diamond63.2\%}$ & $0.227\textcolor{darkpink}{\scriptstyle~\diamond47.1\%}$ & $0.333\textcolor{darkpink}{\scriptstyle~\diamond178.\%}$ & $4.285\textcolor{darkpink}{\scriptstyle~\diamond58.3\%}$ & $0.211\textcolor{darkpink}{\scriptstyle~\diamond35.1\%}$ & $0.350\textcolor{darkpink}{\scriptstyle~\diamond137.\%}$ \\
  \hline   
  \textcolor[rgb]{0.99,0.5,0.0}{EIL} \cite{mai2020erasing} & $1.931\textcolor{darkpink}{\scriptstyle~\diamond20.4\%}$ & $0.285\textcolor{darkpink}{\scriptstyle~\diamond17.2\%}$ &  $0.522\textcolor{darkpink}{\scriptstyle~\diamond77.6\%}$ & $2.167\textcolor{darkpink}{\scriptstyle~\diamond17.5\%}$ & $0.227\textcolor{darkpink}{\scriptstyle~\diamond25.6\%}$ & $0.330\textcolor{darkpink}{\scriptstyle~\diamond151.\%}$ \\
   \textcolor[rgb]{0.99,0.5,0.0}{SPA} \cite{pan2021unveiling} & $5.528\textcolor{darkpink}{\scriptstyle~\diamond72.2\%}$ & $0.221\textcolor{darkpink}{\scriptstyle~\diamond51.1\%}$ & $0.357\textcolor{darkpink}{\scriptstyle~\diamond160.\%}$ & $7.425\textcolor{darkpink}{\scriptstyle~\diamond75.9\%}$ & $0.169\textcolor{darkpink}{\scriptstyle~\diamond68.6\%}$ & $0.262\textcolor{darkpink}{\scriptstyle~\diamond216.\%}$ \\
  \textcolor[rgb]{0.99,0.5,0.0}{TS-CAM} \cite{gao2021ts} & $1.842\textcolor{darkpink}{\scriptstyle~\diamond16.5\%}$ & $0.260\textcolor{darkpink}{\scriptstyle~\diamond28.5\%}$ & $0.336\textcolor{darkpink}{\scriptstyle~\diamond176.\%}$ & $2.104\textcolor{darkpink}{\scriptstyle~\diamond15.1\%}$ & $0.201\textcolor{darkpink}{\scriptstyle~\diamond41.8\%}$ & $0.151\textcolor{darkpink}{\scriptstyle~\diamond449.\%}$  \\
  \hline
  \textcolor[rgb]{0.4,0.0,0.99}{Hotspots} \cite{nagarajan2019grounded} & $1.773\textcolor{darkpink}{\scriptstyle~\diamond13.3\%}$ & $0.278\textcolor{darkpink}{\scriptstyle~\diamond20.1\%}$ & $0.615\textcolor{darkpink}{\scriptstyle~\diamond50.7\%}$ & $1.994\textcolor{darkpink}{\scriptstyle~\diamond10.4\%}$ & $0.237\textcolor{darkpink}{\scriptstyle~\diamond20.3\%}$ & $0.577\textcolor{darkpink}{\scriptstyle~\diamond43.7\%}$\\
  \hline 
  \rowcolor{mygray}
  \textbf{Ours}  & $\bm{1.538}_{\pm 0.017}$ & $\bm{0.334}_{\pm 0.001}$ & $\bm{0.927}_{\pm 0.007}$ & $\bm{1.787}_{\pm0.017}$ & $\bm{0.285}_{\pm 0.002}$ & $\bm{0.829}_{\pm 0.014}$ \\
    \hline
    \Xhline{2.\arrayrulewidth}
    \end{tabular}
  \end{table*}

\subsection{Inference}
\label{TP}
Our test procedure only requires an egocentric object image and an affordance label as input to predict the affordance region. We utilize the class activation mapping \cite{zhou2016learning} by computing a weighted sum of the feature maps $D^i$ of the last convolutional layer to obtain the affordance region heatmap: $Y^{C_a}=\sum_i w^{C_a}_i D^i \label{eq12}$, where $C_a$ is the affordance class, $D^i$ is the $i$-th layer feature map, and $w^{C_a}_i$ is the weight corresponding to the $i$-th neuron under the $C_a$ category.

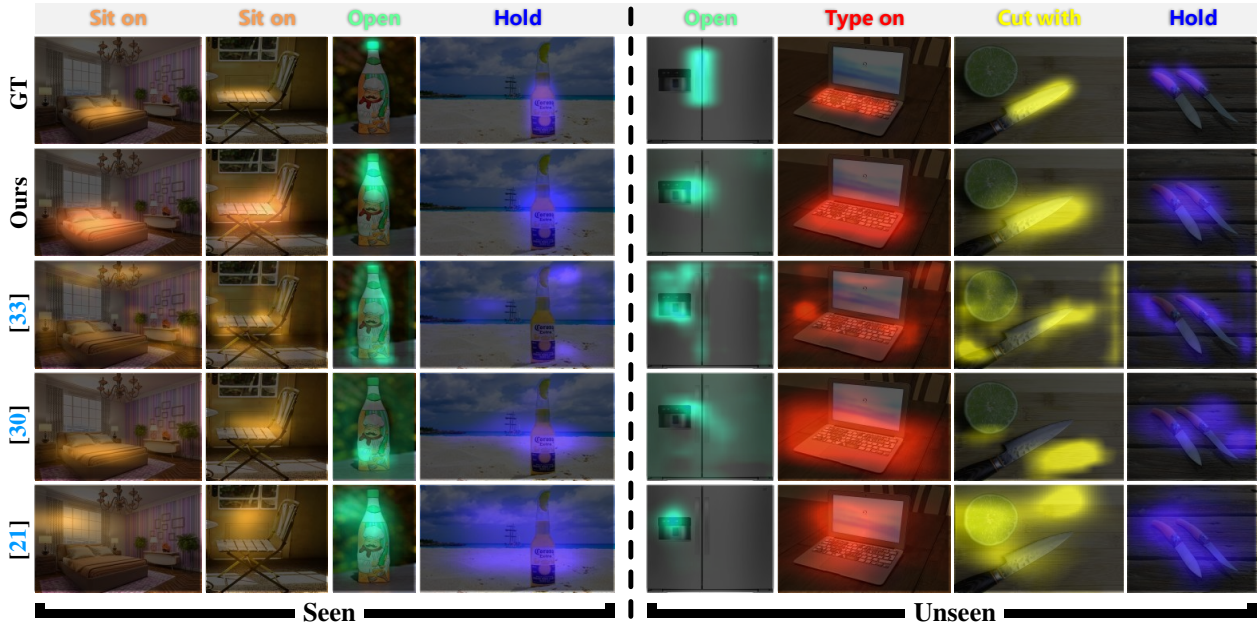
\begin{figure*}[t]
	\centering
		\begin{overpic}[width=0.93\linewidth]{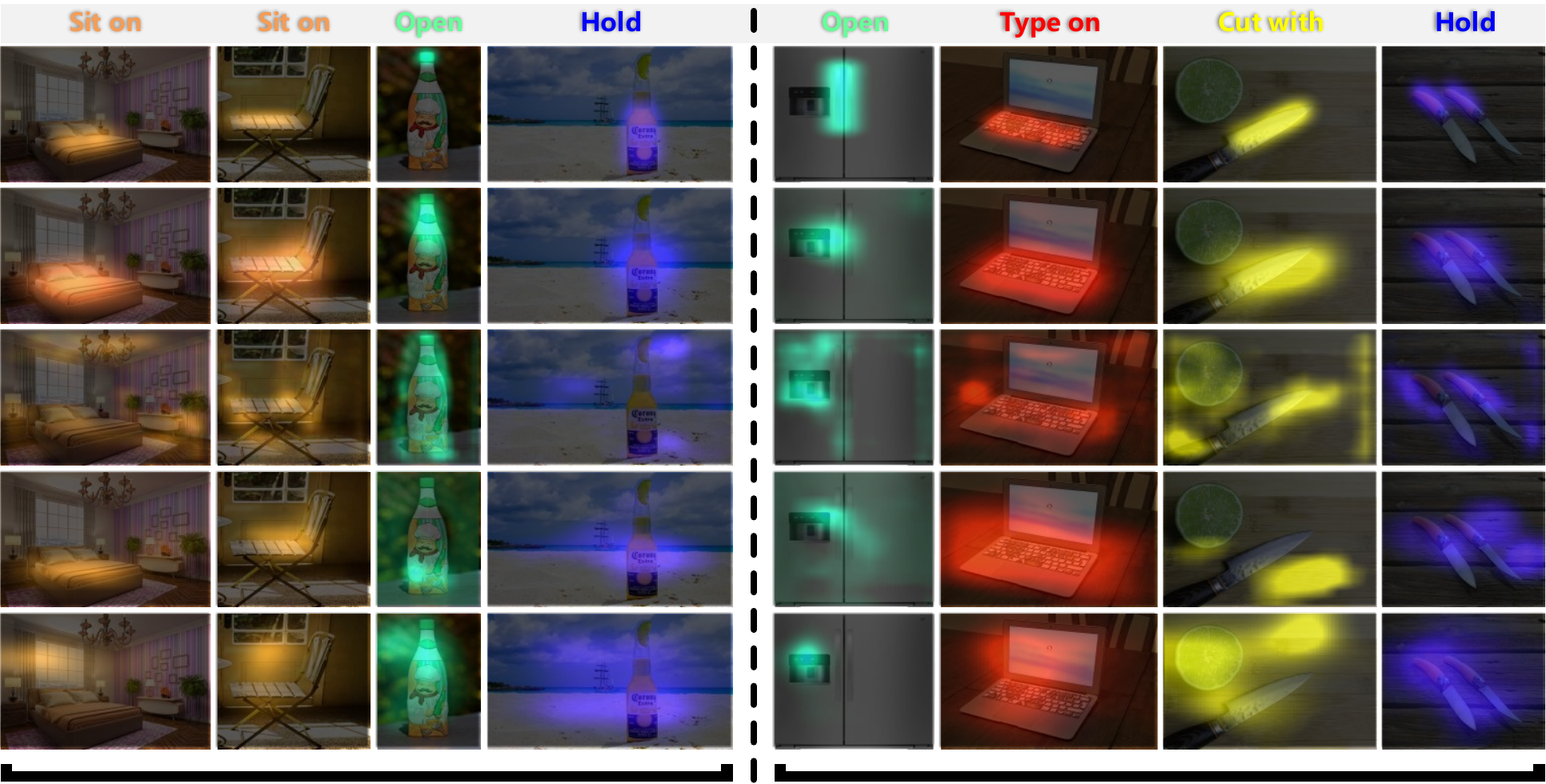}
		\put(-2,41.7){\rotatebox{90}{\small\textbf{GT}}}
		\put(-2.,32){\rotatebox{90}{\small\textbf{Ours}}}
	    \put(-2.1,23.){\rotatebox{90}{\small\textbf{\cite{nagarajan2019grounded}}}}
	    \put(-2.1,13.8){\rotatebox{90}{\small\textbf{\cite{mai2020erasing}}}}
	    \put(-2.1,4.4){\rotatebox{90}{\small\textbf{\cite{kummerer2016deepgaze}}}}
	    \put(21.3,-0.15){\colorbox{white}{{\color{black} \textbf{Seen}}}}
	    \put(71.3,-0.15){\colorbox{white}{{\color{black} \textbf{Unseen}}}}
	\end{overpic}
	\caption{\textbf{Visual affordance heatmaps on the AGD20K dataset.} We select the prediction results of representative methods of affordance grounding (\textcolor[rgb]{0.4,0.0,0.99}{Hotspots} \cite{nagarajan2019grounded}), weakly supervised object localization (\textcolor[rgb]{0.99,0.5,0.0}{EIL} \cite{mai2020erasing}), and saliency detection (\textcolor[rgb]{0.6,0.1,0.1}{DeepGazeII} \cite{kummerer2016deepgaze}) for presentation.}
	\label{result1}
\end{figure*}

\section{Dataset}
\label{sec:dataset}
\textbf{Dataset Collection.\ } The exocentric images are mainly obtained from HICO \cite{chao2018learning} and COCO \cite{lin2014microsoft}. We select images from the HICO dataset according to the verb category and the COCO dataset according to the object category. Then, we manually remove images with ambiguous interactions. To enrich the diversity of the dataset, we further collect $2,112$ exocentric images from free-license websites. Meanwhile, We collect $3,755$ egocentric images from the Internet with free use license according to object categories. 
Some examples are shown in Fig. \ref{dataset} (a). 

\begin{figure*}[t]
    \centering
    \begin{minipage}[t]{0.315\linewidth}
    \centering
    \begin{overpic}[width=0.99\linewidth]{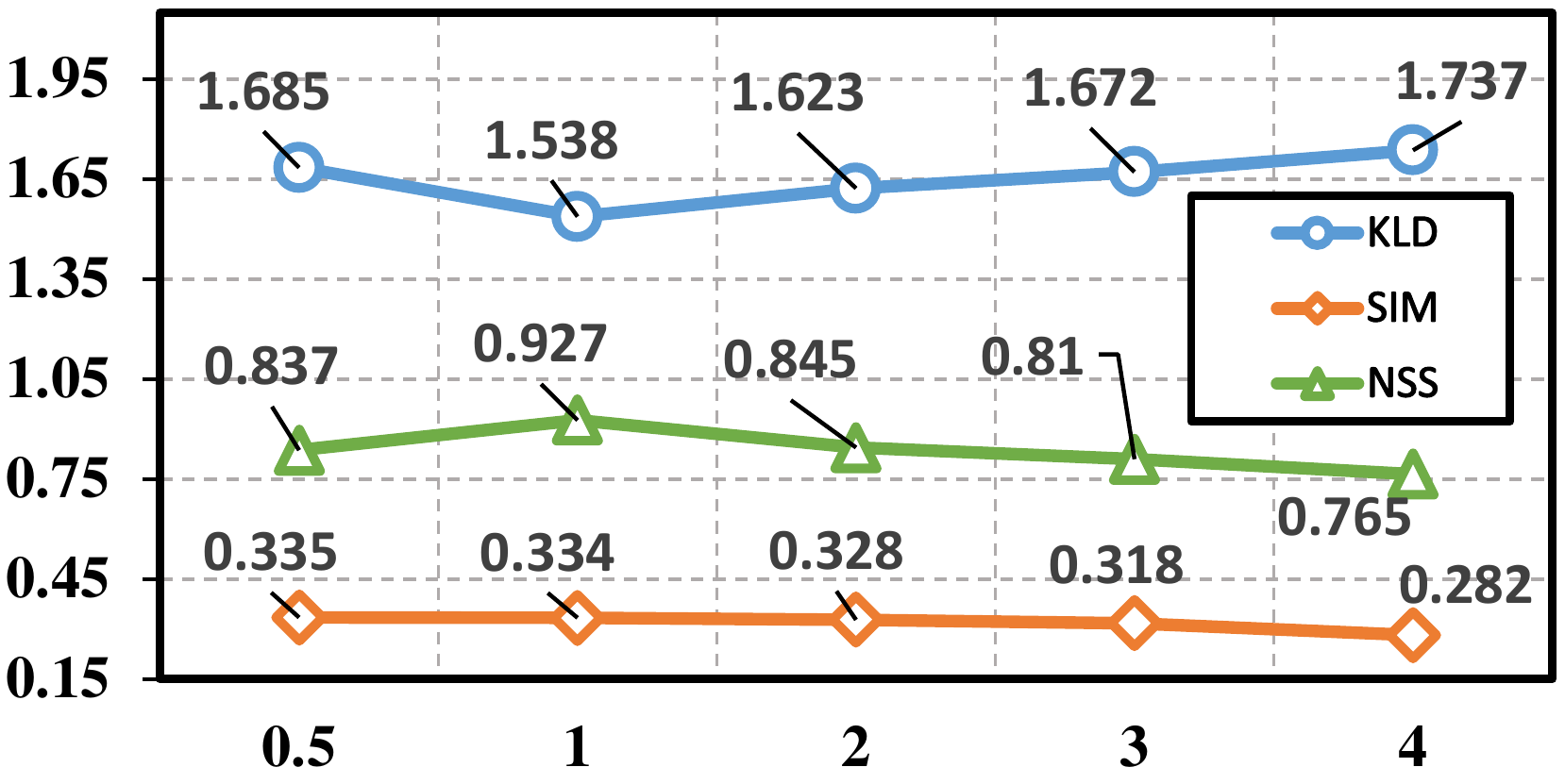}
    \end{overpic}
    \end{minipage}
    \begin{minipage}[t]{0.315\linewidth}
    \centering
    \begin{overpic}[width=0.99\linewidth]{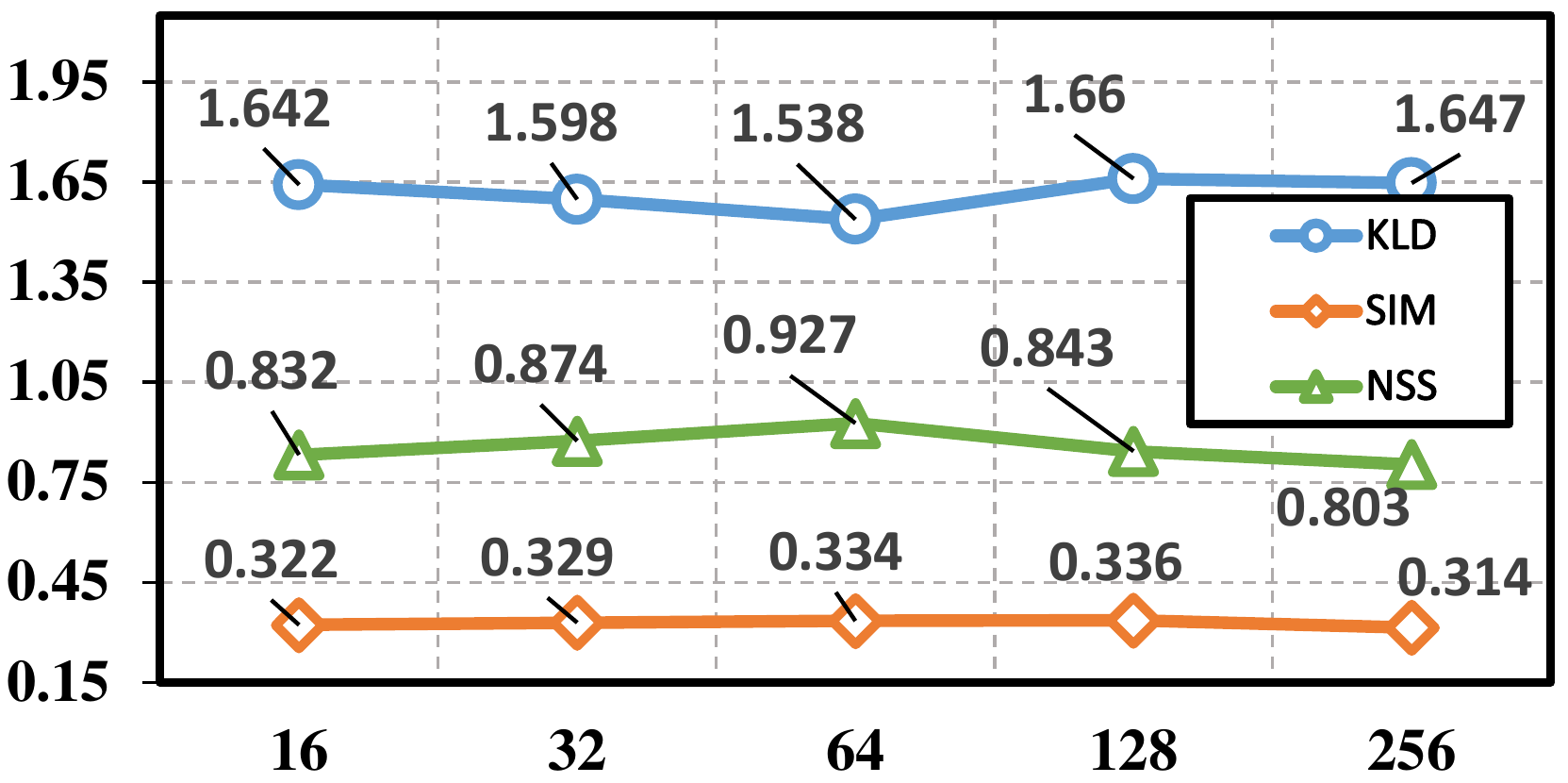}
    \end{overpic}
    \end{minipage}
    \begin{minipage}[t]{0.315\linewidth}
    \centering
    \begin{overpic}[width=0.99\linewidth]{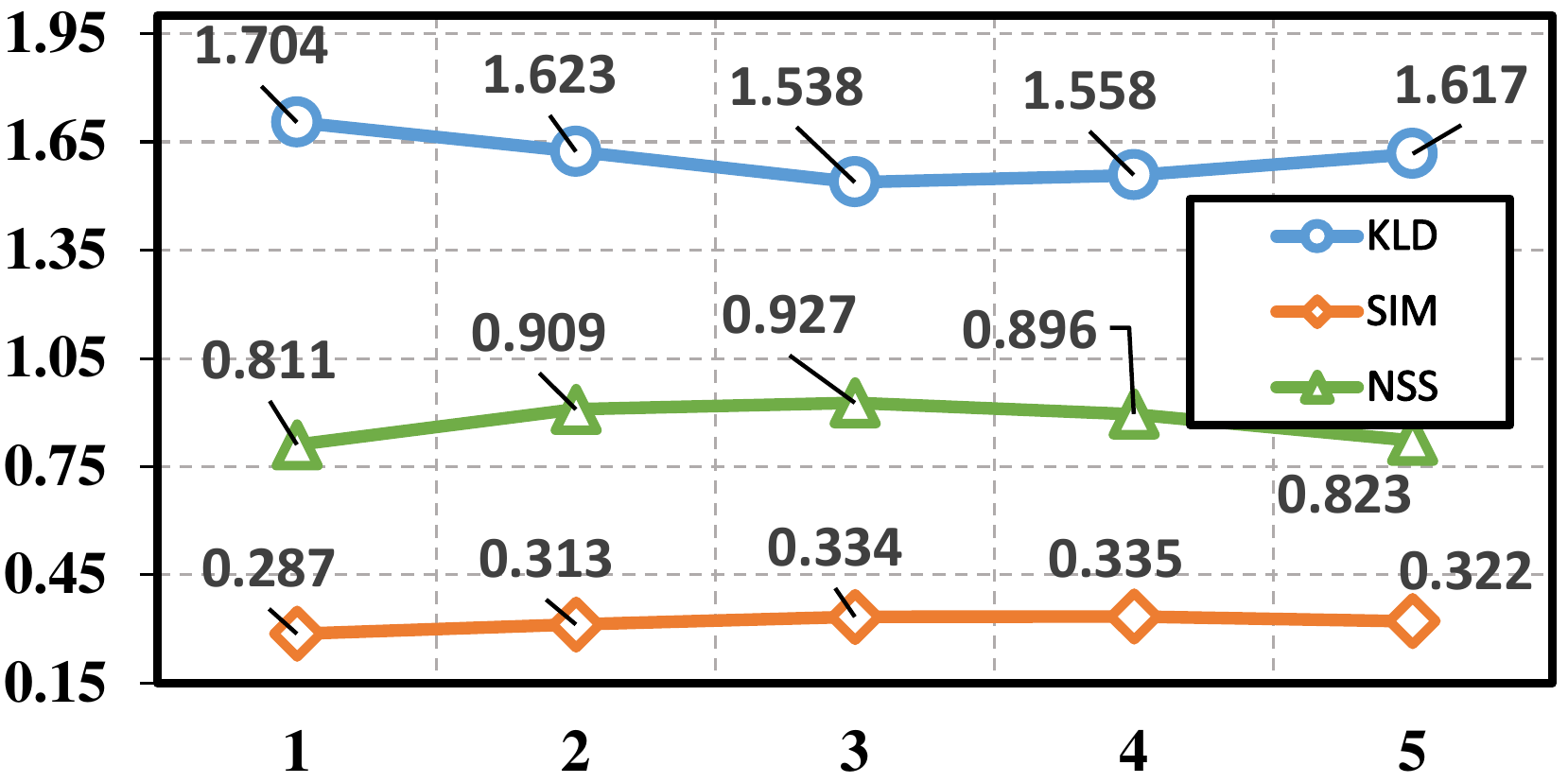}
    \end{overpic}
    \end{minipage}
    \begin{minipage}[t]{0.315\linewidth}
    \centering
    \begin{overpic}[width=0.99\linewidth]{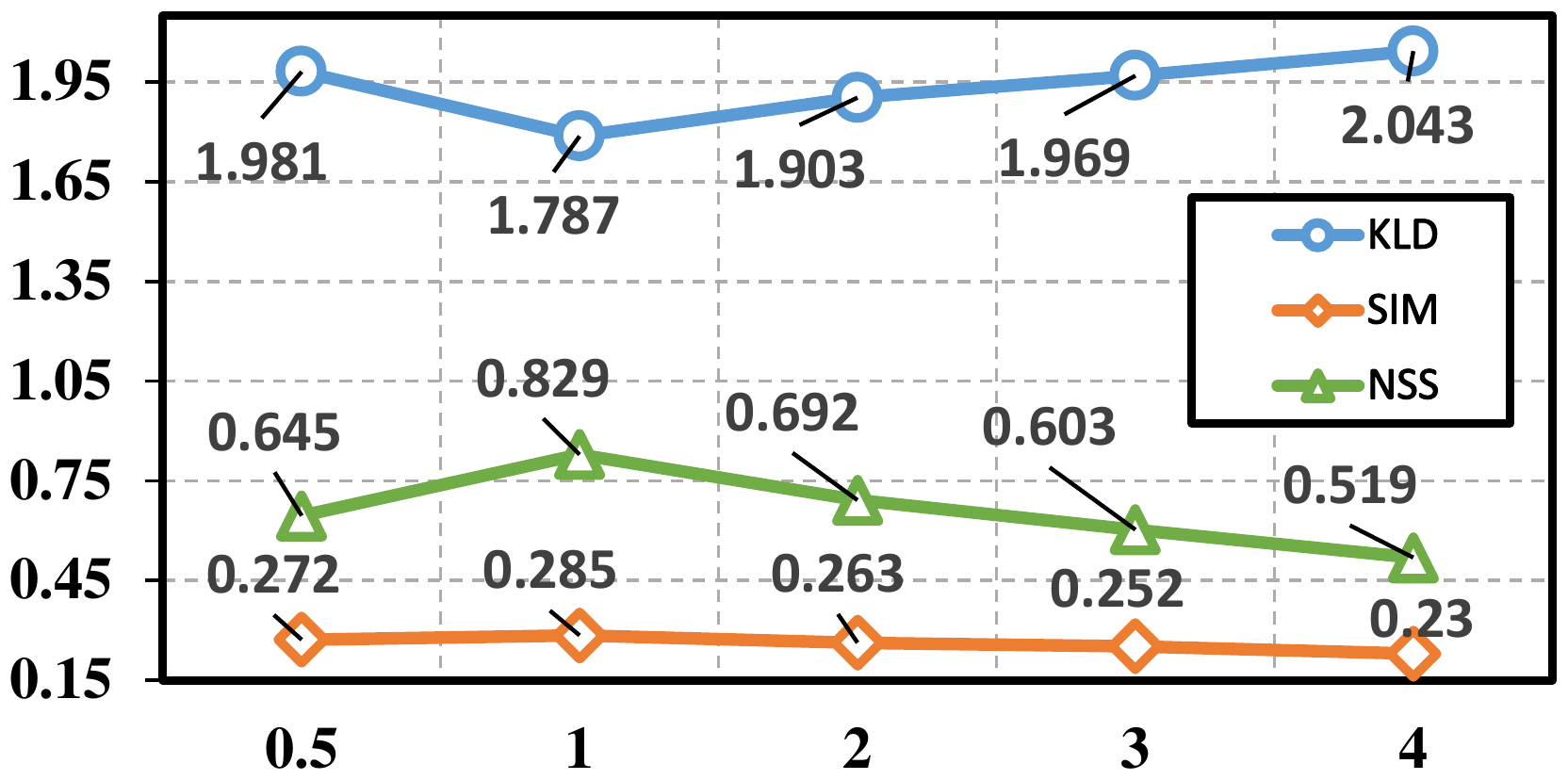}
	    \put(52.7,-4.8){\small$\bm{T}$}
    \end{overpic}
    \end{minipage}
    \begin{minipage}[t]{0.315\linewidth}
    \centering
    \begin{overpic}[width=0.99\linewidth]{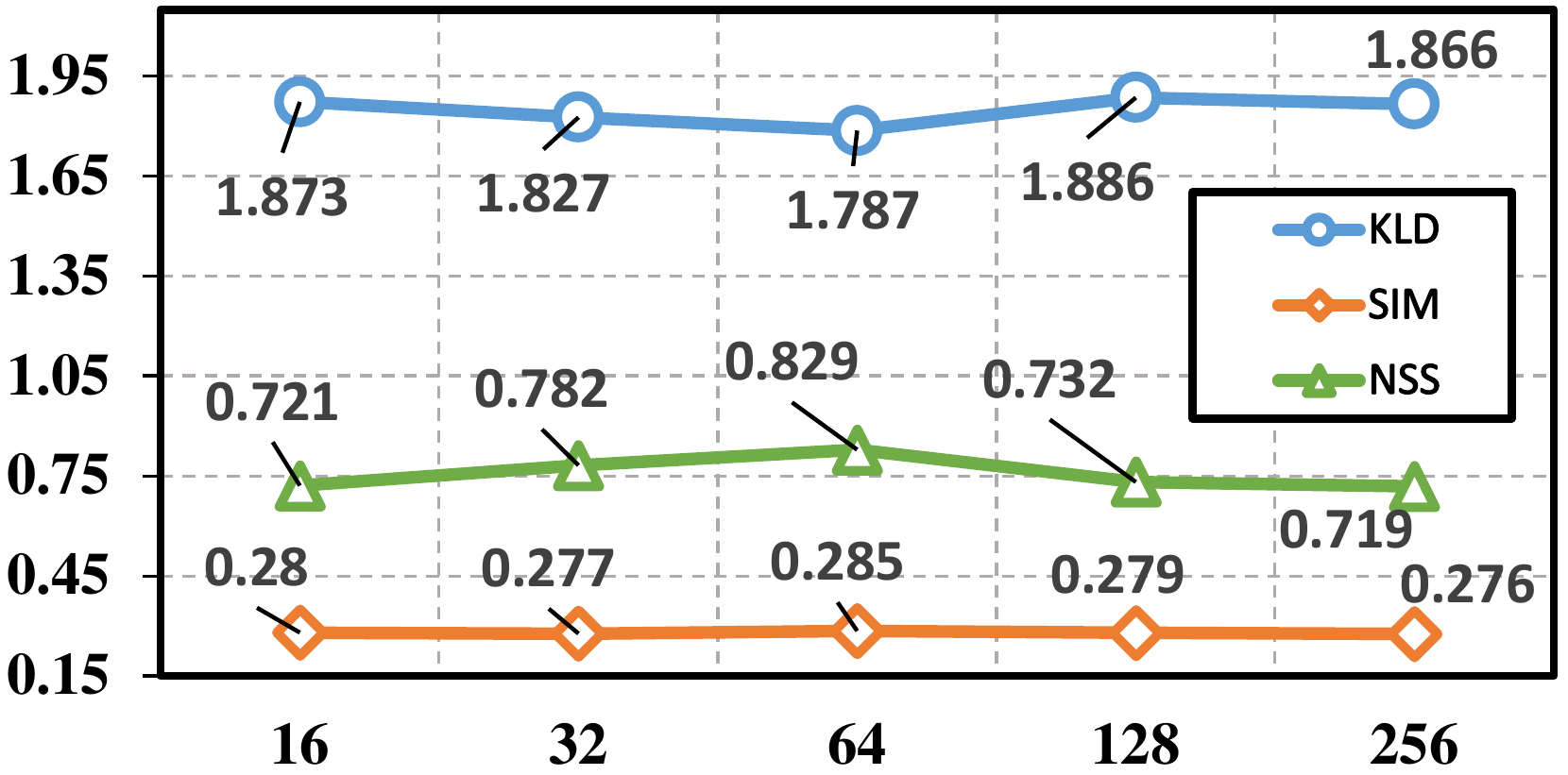}
	    \put(52.5,-4.8){\small{$\bm{r}$}}
    \end{overpic}
    \end{minipage}
    \begin{minipage}[t]{0.315\linewidth}
    \centering
    \begin{overpic}[width=0.99\linewidth]{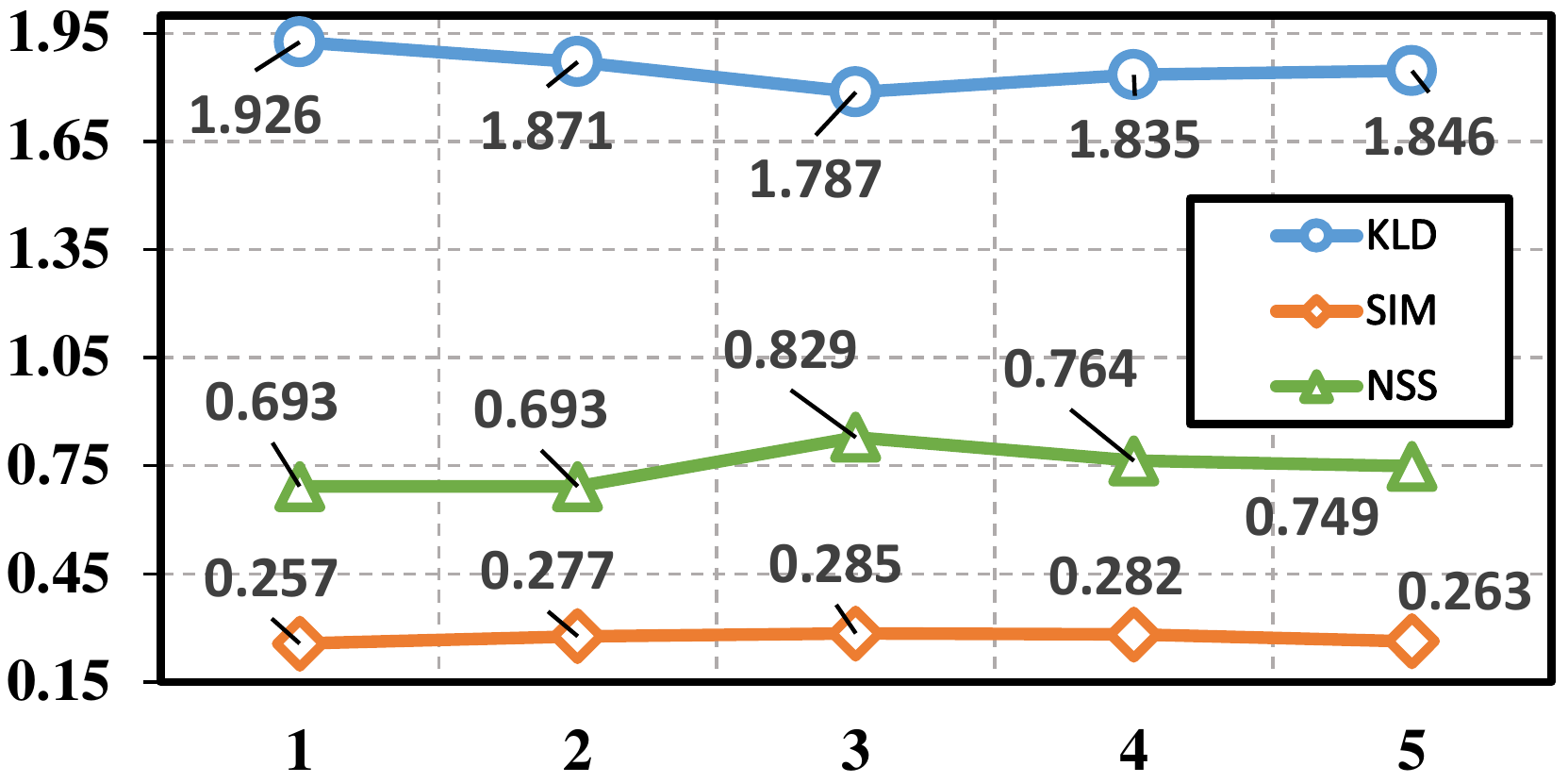}
	    \put(51,-4.8){\small$\bm{N}$}
    \end{overpic}
    \end{minipage}
    \caption{\textbf{Hyper-parameter study.} We investigate the influence of $\boldsymbol{T}$ in the ACP, the rank $r$ of the $W$ in the AIM, and the number of exocentrice images $N$, respectively. The top and bottom columns represent the ``Seen'' and ''Unseen'' experimental settings, respectively.}
    \label{figure:T}
\end{figure*}

\begin{table}[!t]
    \centering
  \footnotesize
  \renewcommand{\arraystretch}{1.}
  \renewcommand{\tabcolsep}{6.5 pt}
   \caption{\textbf{Ablation study.} We investigate the influence of the AIM module, ACP strategy and $L_{KT}$ on model performance.}
   \label{Table:ablation study1}
  \begin{tabular}{c|ccc||ccc}
  \Xhline{2.\arrayrulewidth}
    \hline
  & \textbf{AIM} &  \textbf{ACP} & \bm{$L_{KT}$} & \bm{$\text{KLD} \downarrow$} & \bm{$\text{SIM} \uparrow$} & \bm{$\text{NSS} \uparrow$}  \\
   \hline
   \Xhline{2.\arrayrulewidth}
 \multirow{8}{*}{\rotatebox{90}{\textbf{Seen}}} & & & & $1.985$  & $0.238$ & $0.302$ \\
  & $\checkmark$ & & & $1.750$ & $0.280$ & $0.674$\\
  &   & $\checkmark$ &   & $1.810$ & $0.257$ & $0.687$ \\
   & &  & $\checkmark$ & $1.933$ & $0.241$ & $0.344$ \\
  
 &  $\checkmark$ & $\checkmark$ &  & $1.749$ & $0.286$ & $0.735$ \\
 &  $\checkmark$ &  & $\checkmark$ & $1.664$ & $0.309$ & $0.818$ \\
 &  & $\checkmark$ & $\checkmark$ & $1.741$ & $0.299$ & $0.679$ \\
   \rowcolor{mygray}
  &  $\checkmark$ & $\checkmark$ & $\checkmark$  & \bm{$1.538$} & \bm{$0.334$} & \bm{$0.927$} \\
    \hline
    
    \multirow{8}{*}{\rotatebox{90}{\textbf{Unseen}}} &  & & & $2.059$ & $0.228$ & $0.445$ \\
  & $\checkmark$ & & & $1.933$ & $0.261$ & $0.682$ \\
   
  &   & $\checkmark$ &  & $1.920$ & $0.250$ & $0.666$  \\
  & &  & $\checkmark$ & $1.967$ & $0.265$ & $0.622$ \\
  
 &  $\checkmark$ & $\checkmark$ &  & $1.926$ & $0.269$ & $0.696$ \\
 &  $\checkmark$ &  & $\checkmark$ & $1.916$ & $0.272$ & $0.679$ \\
 &  & $\checkmark$ & $\checkmark$ & $1.922$ & $0.267$ & $0.640$  \\
   \rowcolor{mygray}
  &  $\checkmark$ & $\checkmark$ & $\checkmark$ & \bm{$1.787$} & \bm{$0.285$} & \bm{$0.829$} \\
\hline
    \Xhline{2.\arrayrulewidth}
    \end{tabular}
\end{table}

\textbf{Dataset Annotation.\ } We select $36$ affordance classes commonly used in real-world application scenarios and assign labels to each image based on the interaction between human and object in each exocentric image. Given the object class contained in each affordance class, we assign affordance labels based on the object class in the egocentric images. The testing process requires pixel-level labels to calculate objective metrics. As the annotation approach in \cite{fang2018demo2vec}, we take the form of points for regions of interaction, in which the dense points are for regions of frequent interaction and vice versa. Then, heatmaps of affordance regions can be obtained from the points as \cite{fang2018demo2vec}. Some annotation examples are shown in Fig. \ref{dataset} (a).

\textbf{Statistic Analysis.\ } To obtain deeper insights into our AGD20K dataset, we show its important features from the following aspects. The distribution of categories in the dataset is shown in Fig. \ref{dataset} (b), which shows that the dataset contains a wide range of affordance/object categories in diverse scenarios. The affordance word cloud is shown in Fig. \ref{dataset} (c). The confusion matrix of affordance and object categories is shown in Fig. \ref{dataset} (d). It shows a multi-to-multi relationship between affordance and object categories, posing a significant challenge for the affordance grounding task. See supplementary materials for more details.

\section{Experiments}
\subsection{Benchmark Setting}
\label{5.1}
To provide a comprehensive evaluation, we choose three commonly used metrics  \textbf{K}ullback-\textbf{L}eibler \textbf{D}ivergence (\textbf{KLD}) \cite{bylinskii2018different}, \textbf{SIM}ilarity (\textbf{SIM}) \cite{swain1991color} and \textbf{N}ormalized \textbf{S}canpath \textbf{S}aliency (\textbf{NSS}) \cite{peters2005components}, see supplementary material for details of each metric.  Our model is implemented in PyTorch and trained with the SGD optimizer. The input images are randomly clipped from $256 \times 256$ to $224 \times 224$ with random horizontal flipping. 
We train the model for $35$ epochs on a single NVIDIA $1080$ti GPU with an initial learning rate of $1e$-$3$. 
The hyper-parameters $\lambda_1$, $\lambda_2$ and $\lambda_3$ are set to $1$, $0.5$ and $0.5$ respectively. The hyper-parameter $\boldsymbol{T}$ in the ACP is set to $1$. The rank $r$ of the dictionary matrix $W$ and the number of iterations in the AIM are set to $64$ and $6$ respectively. The number of exocentric images $N$ is set to $3$. Besides, three saliency detection models (\textcolor[rgb]{0.6,0.1,0.1}{Mlnet} \cite{cornia2016deep}, \textcolor[rgb]{0.6,0.1,0.1}{DeepGazeII} \cite{kummerer2016deepgaze}, \textcolor[rgb]{0.6,0.1,0.1}{EgoGaze} \cite{huang2018predicting}), three weakly supervised object localization models (\textcolor[rgb]{0.99,0.5,0.0}{EIL} \cite{mai2020erasing}, \textcolor[rgb]{0.99,0.5,0.0}{SPA} \cite{pan2021unveiling}, \textcolor[rgb]{0.99,0.5,0.0}{TS-CAM} \cite{gao2021ts}) and one affordance grounding model (\textcolor[rgb]{0.4,0.0,0.99}{Hotspots} \cite{nagarajan2019grounded}) are chosen for comparison. We design two different settings: 1) ``Seen'', \ie, the training set and the test set contain the same object categories, and 2) ``Unseen'', \ie, the object categories in the training set and the test set do not overlap.

\subsection{Quantitative and Qualitative Comparisons}
\label{5.2}
The experimental results are shown in Table \ref{Table:1}. Our method achieves the best results in both ``Seen'' and ``Unseen'' settings. Taking KLD as the metric, our method improves 17.2\% compared to the best saliency model, 16.5\% over the best weakly supervised object localization (WSOL) model, and 13.3\% over the affordance grounding model in the ``Seen'' setting. Our method with the ``Unseen'' setting improves 10.2\% compared to the best saliency model, surpasses the best WSOL model by 15.1\%, and exceeds the affordance grounding model by 10.4\%. It indicates that our method can effectively transfer the affordance knowledge from the exocentric view to the object in the egocentric view and has a good generalization ability for unseen objects. 

In addition, we visualize the affordance maps in ``Seen'' and ``Unseen'' settings, as shown in Fig. \ref{result1}. It shows that our method can obtain more accurate prediction results for affordance grounding. While ``Sit on'' contains objects with different appearances (``bed'' and ``chair''), our method can capture the common features of the affordance region and obtain better prediction results. Since ``bottle'' has two different affordances, ``Open'' and ``Hold'', the network predicts different affordance regions. In the ``Unseen'' setting, the ``knife'' has two different affordances, ``Hold'' and ``Cut with''. Our method can locate different affordance regions based on the learned affordance knowledge, demonstrating its superior generalization capability. 

\begin{table}[!t]
 \centering
  \footnotesize
  \renewcommand{\arraystretch}{1.}
  \renewcommand{\tabcolsep}{3.pt}
   \caption{{\textbf{Different classes.}
   The KLD results of different methods on some representative affordance categories.}}
   \label{Table:class}
  \begin{tabular}{r||ccccc}
    \hline
    \Xhline{2.\arrayrulewidth}
   \hline
   \textbf{Classes}  & \textbf{Hold} & \textbf{Swing} & \textbf{Drink with} & \textbf{Lie on} & \textbf{Brush with}
\\   \hline
\Xhline{2.\arrayrulewidth}
  \textcolor[rgb]{0.6,0.1,0.1}{Mlnet} \cite{cornia2016deep}   & $6.762$ & $9.248$ & $4.497$ & $4.767$ & $6.215$  \\
  \textcolor[rgb]{0.6,0.1,0.1}{DeepGazeII} \cite{kummerer2016deepgaze}  & $2.071$ & $2.478$ & $2.067$ & $1.602$ & $2.385$ \\
  \textcolor[rgb]{0.6,0.1,0.1}{EgoGaze} \cite{huang2018predicting}  & $4.671$ & $6.723$ & $4.268$ & $2.921$ & $5.135$ \\
  \hline   
  \textcolor[rgb]{0.99,0.5,0.0}{EIL} \cite{mai2020erasing}  & $2.008$ & $2.486$ & $2.254$ & $1.377$ & $3.003$ \\

   \textcolor[rgb]{0.99,0.5,0.0}{SPA} \cite{pan2021unveiling}  & $3.006$ & $6.720$ & $7.683$ & $4.006$ & $8.043$  \\
  
  \textcolor[rgb]{0.99,0.5,0.0}{TS-CAM} \cite{gao2021ts}  & $1.628$ & $2.420$ & $2.300$ & $1.370$ & $2.642$  \\
  \hline
  \textcolor[rgb]{0.4,0.0,0.99}{Hotspots} \cite{nagarajan2019grounded}  & $1.770$ & $2.178$ & $1.942$ & $1.566$ & $2.154$ \\
  \hline 
  \rowcolor{mygray}
  \textbf{Ours}  & \bm{$1.594$} & \bm{$2.161$} & \bm{$1.748$} & \bm{$1.039$} & \bm{$2.040$}  \\
    \hline
    \Xhline{2.\arrayrulewidth}
    \end{tabular}
  \end{table}

\begin{figure}[t]
    \centering
    \begin{minipage}[t]{0.39\linewidth}
    \centering
    \begin{overpic}[width=0.99\linewidth]{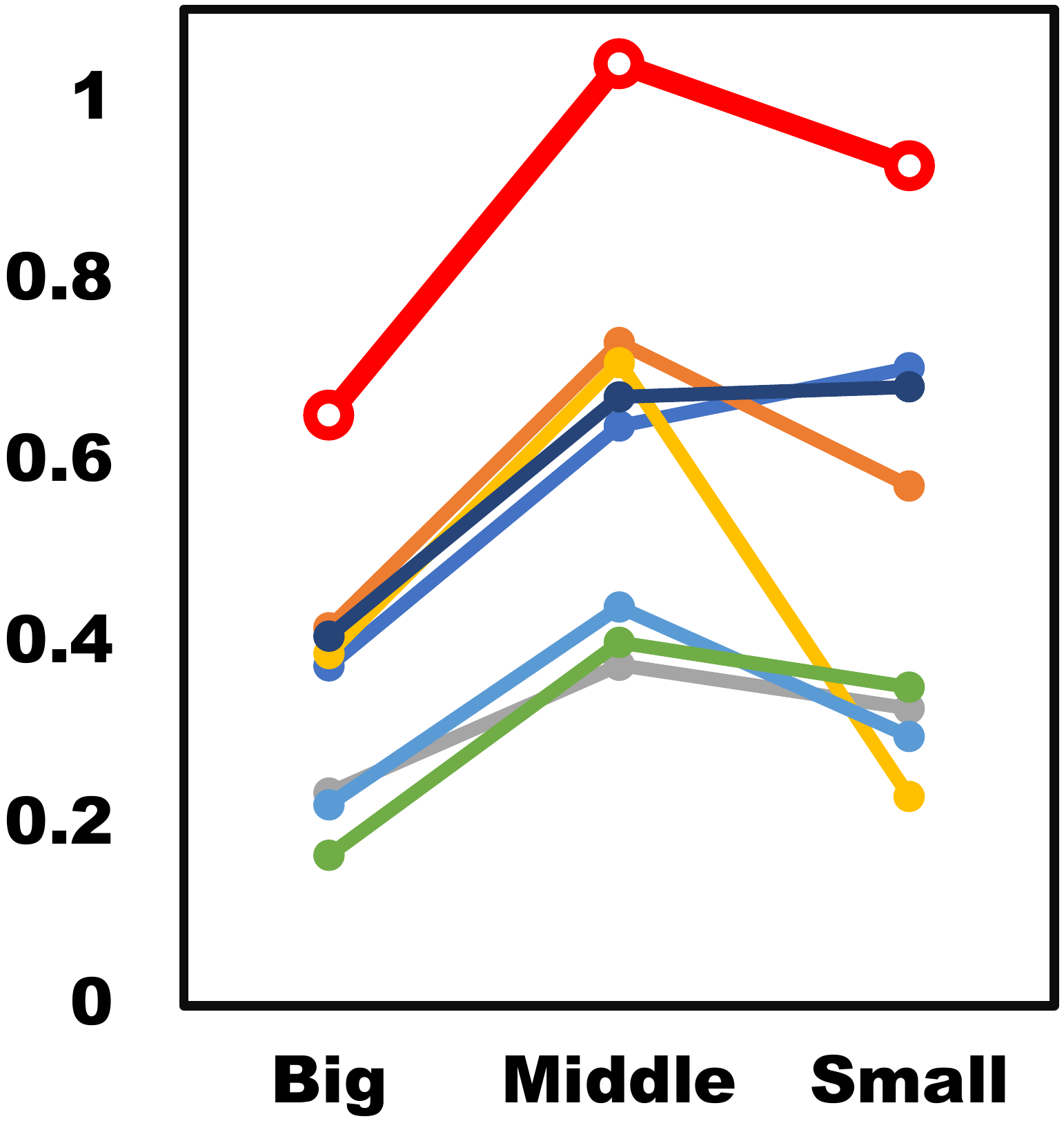}
    \put(46,-6.9){\small\textbf{Seen}}
    \end{overpic}
    \end{minipage}
    \begin{minipage}[t]{0.56\linewidth}
    \centering
    \begin{overpic}[width=0.99\linewidth]{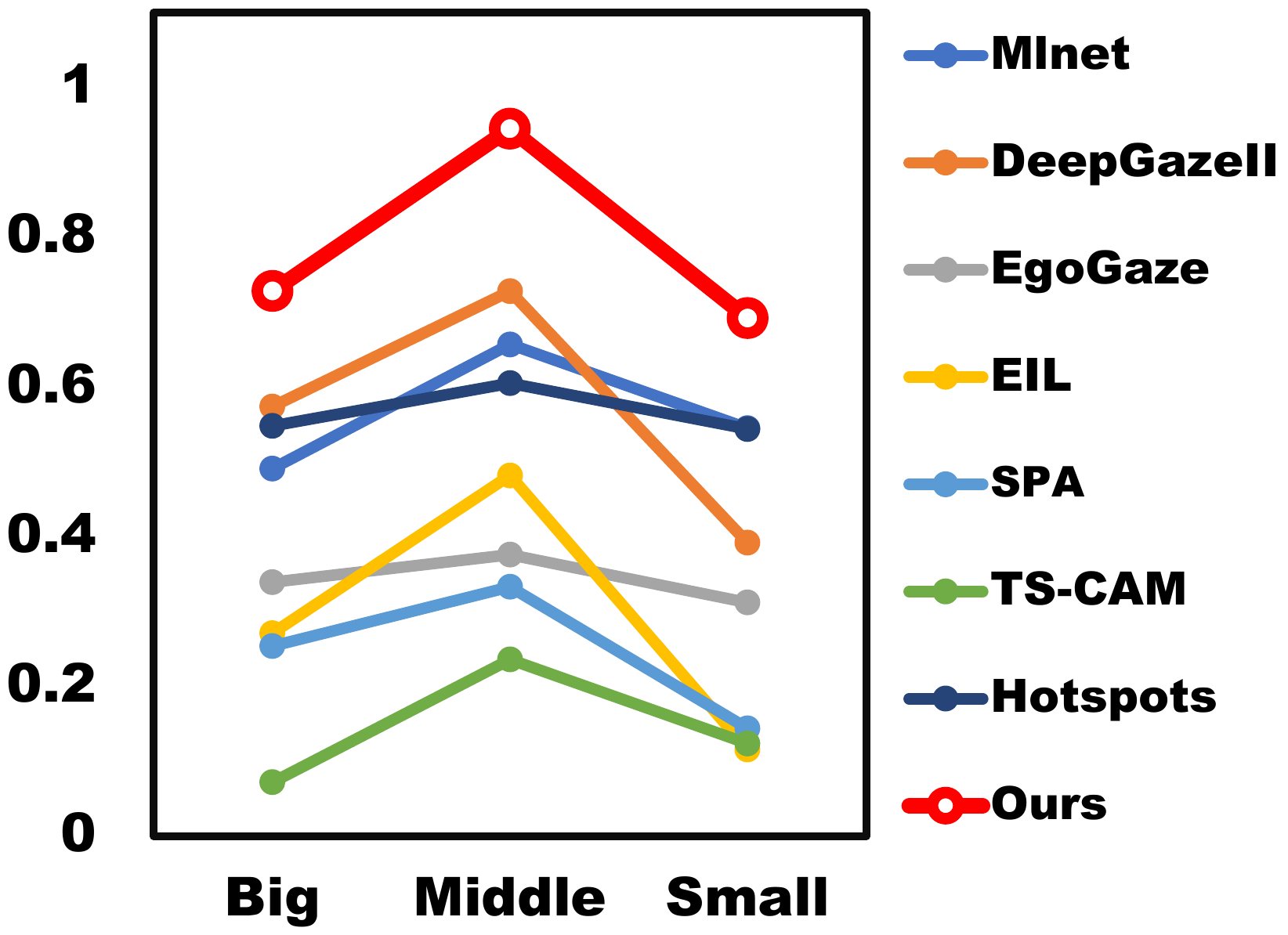}
    \put(29,-5.1){\small\textbf{Unseen}}
    \end{overpic}
    \end{minipage}
    \caption{\textbf{Different scales.} We split the test set into three subsets of ``Big'', ``Middle'' and ``Small'' according to the ratio of mask to the whole image, and show the results of the NSS metrics.}
    \label{figure:scale}
\end{figure}

\subsection{Ablation Study}
\label{5.3}
The ablation study results are shown in Table \ref{Table:ablation study1}. It confirms the ability of the AIM module to learn affordance-specific features from diverse exocentric interactions, which play a significant role in improving network performance. The ACP strategy improves more obviously than $L_{KT}$, indicating that the preservation of affordance co-relation can more effectively improve the network's ability to perceive and locate affordance regions. In addition, we investigate the influence of different hyper-parameter settings of $\boldsymbol{T}$ in the ACP strategy (shown in Fig. \ref{figure:T} (left)), the rank $r$ of the dictionary matrix $W$ in the AIM module (shown in Fig. \ref{figure:T} (middle)), and different exocentric images $N$(see Fig. \ref{figure:T} (right)). It can be seen that the performance of the model is more influenced by $\boldsymbol{T}$, while the rank $r$ does not have a significant impact on the results. The number of exocentric images taken from $2$ to $3$ has a larger impact on the model. For $N = 1$, our model still outperforms most contenders.

\begin{table}[!t]
  \centering
  \renewcommand{\arraystretch}{1.}
  \renewcommand{\tabcolsep}{5pt}
   \small
   \caption{\textbf{Different sources.} ``Exo'' means that training only uses exocentric images, ``Exo\&Ego'' means that training uses both exocentric and egocentric images.}
   \label{Table:different sources}
  \begin{tabular}{c|c|c||ccc}
    \hline
    \Xhline{2.\arrayrulewidth}
  &   \textbf{Method} & \textbf{Source} & $\text{KLD} \downarrow$ & $\text{SIM} \uparrow$ & $\text{NSS} \uparrow$ \\
   \hline
   \Xhline{2.\arrayrulewidth}
 \multirow{7}{*}{\rotatebox{90}{\textbf{Seen}}} & 
   \multirow{2}{*}{ EIL \cite{mai2020erasing}}  & Exo  &  $1.931$ & $0.285$ & $0.522$ \\
  &     & Exo\&Ego & $2.156$ & $0.321$ & $0.747$ \\
  \cline{2-6}
  & \multirow{2}{*}{SPA \cite{pan2021unveiling}}  & Exo & $5.528$ & $0.221$ & $0.357$  \\
    
  &   & Exo\&Ego & $4.312$ & $0.252$ & $0.494$  \\
  \cline{2-6}
  & \multirow{2}{*}{TS-CAM \cite{gao2021ts}}  & Exo & $1.842$ & $0.260$ & $0.336$ \\
  &   & Exo\&Ego & $1.707$ & $0.290$ & $0.622$  \\
 \cline{2-6}
 \rowcolor{mygray}
 & \textbf{Ours} & Exo\&Ego & \bm{$1.538$} & \bm{$0.334$} & \bm{$0.927$}  \\
    \hline
    \Xhline{2.\arrayrulewidth}
    \multirow{7}{*}{\rotatebox{90}{\textbf{Unseen}}} &  \multirow{2}{*}{EIL \cite{mai2020erasing}}  & Exo & $2.167$ & $0.227$ & $0.330$ \\
  &     & Exo\&Ego & $2.029$ & $0.256$ & $0.529$ \\
  \cline{2-6}
  & \multirow{2}{*}{SPA \cite{pan2021unveiling}}  & Exo & $7.425$ & $0.169$ & $0.262$  \\
    
  &   & Exo\&Ego & $6.174$ & $0.209$ & $0.433$   \\
  \cline{2-6}
  & \multirow{2}{*}{TS-CAM \cite{gao2021ts}}  & Exo & $2.104$ & $0.201$ & $0.151$  \\
  &  & Exo\&Ego & $2.002$ & $0.228$ & $0.305$  \\
  \cline{2-6}
  \rowcolor{mygray}
  & \textbf{Ours} & Exo\&Ego & $\bm{1.787}$ & $\bm{0.285}$ & $\bm{0.829}$   \\
 \hline
    \Xhline{2.\arrayrulewidth}
    \end{tabular}
\end{table}

\subsection{Performance Analysis}
\label{6.1}
\textbf{Different Classes.\ } The KLD metrics on some representative categories are shown in Table \ref{Table:class}. ``Hold'' and ``Swing'' both contain diverse object categories with different appearances. ``Drink with'' and ``Hold'' contain overlapped object categories but have completely different affordance regions. Objects of ``Lie on'' are generally labeled with a larger region, while  those of ``Brush with'' are generally smaller. Our model exceeds others regarding different aspects of the challenge, which confirms its robustness. See supplementary material for the KLD metrics for each category.

\textbf{Different Scales.\ } We divide the test set into ``big'', ``middle'' and ``small'' splits according to the proportion of mask to the whole image (see supplementary material for details). The test results are shown in Fig. \ref{figure:scale}. Our model  outperforms all other methods in all splits on both settings, showing its ability to capture the intrinsic affordance properties of objects, even in more challenging cases. The performance of the experimental results on all metrics are shown in the supplementary material.

\textbf{Different Sources.\ } The results for different sources are shown in Table \ref{Table:different sources}. It shows that using both exocentric and egocentric images improves most methods, but the improvement is limited. Our method still surpasses all models,  showing that the knowledge transfer from explicitly cross-views is effective in learning from exocentric diverse interactions to egocentric invariant affordance representation. The performance of the experimental results on all metrics are shown in the supplementary material.

\textbf{Limitations.\ }
Our method still has limitations, \eg, the predicted affordance maps may contain intermediate background regions when multiple objects appear and irrelevant background regions may be activated for slender object. In the future, we will refer to \cite{pan2021unveiling} to refine the generated results to obtain more accurate results.

\section{Conclusion}
\label{6}
In this paper, we make an attempt to address a new challenging task named affordance grounding from exocentric view. Specifically, we propose a novel cross-view knowledge transfer framework that can extract invariant affordance from diverse exocentric interactions and transfer it to egocentric view. We establish a large affordance grounding dataset named AGD20K, which contains 20K well-annotated images, serving as a pioneer testbed for the task. Our model outperforms representative models from related areas and can serve as a strong baseline for future research. 

\textbf{Broader Impacts.\ }
The research on affordance grounding from exocentric view will advance the realization of embodied intelligence. However, harmful human demonstrations (risky behaviors) may lead to negative guidance for the agent, which should be prohibited by strict legislation.

\textbf{Acknowledgments.}This work was supported by National Key R\&D Program of China under Grant 2020AAA0105701, National Natural Science Foundation of China (NSFC) under Grants 61872327 and ARC FL-170100117.

{\small
\bibliographystyle{ieee_fullname}
\bibliography{cvpr}
}

\end{document}